\documentclass[]{sn-jnl}
\pdfoutput=1

\usepackage[T1]{fontenc}
\usepackage[utf8]{inputenc}
\usepackage{bigfoot}
\usepackage{color}
\usepackage{babel}
\usepackage{array}
\usepackage{rotfloat}
\usepackage{multirow}
\usepackage{amsmath}
\usepackage{amsthm}
\usepackage{amssymb}
\usepackage{graphicx}
\usepackage[numbers,sort&compress]{natbib}
\usepackage{multibib}
\newcites{searchresults}%
     {Search Results} 

\usepackage{setspace}
\usepackage{caption}
\newcommand{\rev}[1]{\textcolor{black}{#1}}

\begin{document}

\title[Integrated Planning in Hospitals: A Review]{Integrated Planning in Hospitals: A Review}

\author[1,2]{\fnm{Sebastian} \sur{Rachuba}}\email{s.rachuba@utwente.nl}

\author[3]{\fnm{Melanie} \sur{Reuter-Oppermann}}\email{melanie.reuter-oppermann@dgre.org}

\author*[4,5]{\fnm{Clemens} \sur{Thielen}}\email{clemens.thielen@tum.de}

\affil[1]{\orgdiv{Center for Healthcare Operations Improvement \& Research}, \orgname{University of Twente}, \orgaddress{\street{Drienerlolaan 5}, \city{Enschede}, \postcode{7522NB}, \country{The~Netherlands}}}

\affil[2]{\orgdiv{Medical School}, \orgname{University of Exeter}, \orgaddress{\street{Heavitree Road}, \city{Exeter}, \postcode{EX1~2LU}, \country{United Kingdom}}}

\affil[3]{\orgdiv{Department of Health, Care and Public Health Research Institute (CAPHRI)}, \orgname{Maastricht University}, \orgaddress{\street{Universiteitssingel 40}, \postcode{6229 ER}~\city{Maastricht}, \country{Netherlands}}}

\affil[4]{\orgdiv{TUM Campus Straubing for Biotechnology and Sustainability}, \orgname{Weihenstephan-Triesdorf University of Applied Sciences}, \orgaddress{\street{Am~Essigberg 3}, \postcode{94315}~\city{Straubing}, \country{Germany}}}

\affil[5]{\orgdiv{Department of Mathematics, School of Computation, Information and Technology}, \orgname{Technical University of Munich}, \orgaddress{\street{Boltzmannstr.~3}, \postcode{85748}~\city{Garching bei München}, \country{Germany}}}

\abstract{Efficient planning of scarce resources in hospitals is a challenging task for which a large variety of Operations Research and Management Science approaches have been developed since the 1950s. While efficient planning of single resources such as operating rooms, beds, or specific types of staff can already lead to enormous efficiency gains, \emph{integrated planning} of several resources has been shown to hold even greater potential, and a large number of integrated planning approaches have been presented in the literature over the past decades. 
This paper provides the first literature review that focuses specifically on the Operations Research and Management Science literature related to integrated planning of different resources in hospitals. We collect the relevant literature and analyze it regarding different aspects such as uncertainty modeling and the use of real-life data. Several cross comparisons reveal interesting insights concerning, e.g., relations between the modeling and solution methods used and the practical implementation of the approaches developed. Moreover, we provide a high-level taxonomy for classifying different resource-focused integration approaches and point out gaps in the literature as well as promising directions for future research.}


\maketitle

\section{Introduction}
A well-performing healthcare system is a crucial part of a modern society and determines people's lives and livelihood~\cite{WHO2000}. The importance of a healthcare system is also reflected in the enormous spending required.  For instance, an unprecedented 10.9\% of the GDP of the European Union was devoted to healthcare in 2020~\cite{OECD:health-at-a-glance}.

It is widely recognized that demand for healthcare will further increase in the future due to demographic changes such as growth in elderly population in nearly all developed countries and increased longevity~\cite{Rais+Viana:healt-care-or-survey}. For instance, the share of \rev{people over 65 (over 80)} in Germany increased from 20.6\,\% (5.2\,\%) to 22.0\,\% (7.1\,\%) between 2011 and 2021~\cite{OECD:webpage}. Due to unavailability of crucial resources such as staff (particularly physicians~\cite{Bodenheimer+Smith:primary-care} and nurses~\cite{Aiken+etal:nurse-burnout}), however, increased demand cannot be addressed by simply increasing healthcare spending to fund additional treatment capacities. Instead, the available scarce resources have to be used as efficiently as possible in order to ensure the continued provision of high-quality care in the healthcare sector.

Good planning for efficient resource use in healthcare is a very challenging task due to various inherent characteristics that complicate planning decisions on all hierarchical levels -- from long-term or strategic planning down to operational online decision making. These characteristics include (1) the wide-spread organizational subdivision of central entities such as hospitals~\cite{Roth+vanDierdonck:hospital-resource-planning,Porter+Teisberg:physicians-change-future}, (2) conflicting objectives and lack of cooperation between involved parties such as physicians, nurses, or administrators~\cite{Hans:framework}, (3) unavailability of crucial information required for planning and control~\cite{Carter:mismanagement}, and (4) uncertainty and high fluctuation in the daily requirements for care~\cite{Erhard+etal:phys-sched-survey}. Consequently, advanced planning methods are necessary in order to provide high-quality support to decision makers and use the available resources efficiently. 

Operations Research (OR) and Management Science (MS) offer a variety of scientific approaches for the efficient management and planning of limited resources that are applied with enormous success in healthcare since the 1950s~\cite{Hulshof:taxonomy}. Extensive overviews on OR/MS in healthcare are provided by Pierskalla and Brailer~\cite{Pierskalla:survey}, Rais and Viana~\cite{Rais+Viana:healt-care-or-survey}, Hulshof et al.~\cite{Hulshof:taxonomy}, and Jha et al.~\cite{Jha:survey}. Surveys focused on methods for a particular, important resource are available for operating rooms~\cite{guerriero2011:survey-or,vanriet2015:trade-offs-or}, inpatient beds~\cite{He2019:sys-review-bed-management}, intensive care units~\cite{Bai+etal:intensive-care-review}, physicians~\cite{Erhard+etal:phys-sched-survey}, and nurses~\cite{Burke+etal:survey,Causmaecker:nurse-rostering-categorisation,Benazzouz+etal:survey}.

An efficient planning of single resources such as operating rooms, beds, or specific types of staff can already lead to enormous efficiency gains and improved resource utilization in a healthcare system. Approaches that focus on isolated decision making in this way, however, ignore the inherent complex interactions between different resources or organizational units~\cite{Hulshof:taxonomy} and, therefore, often lead to suboptimal decisions on a system level. This is particularly apparent in hospitals, which collect large amounts of advanced technology and clinical specialization, but are usually subdivided into a variety of autonomously managed departments~\cite{Roth+vanDierdonck:hospital-resource-planning,Porter+Teisberg:physicians-change-future,Hulshof:PHD}. Consequently, a need for OR/MS models that focus on \emph{integrated planning} of several resources has been identified~\cite{Jun+etal:discrete-event-survey,vanberkelSurvey,Hulshof:taxonomy}. This \emph{vertical integration} (integration across different resources) is considered to show great potential, and an increase in publications presenting vertically integrated approaches has been observed~\cite{Cardoen:operating-room-survey,vanberkelSurvey,Hulshof:taxonomy}. It complements \emph{horizontal integration}, which refers to integration across different hierarchical, or temporal, decision making levels, which are traditionally subdivided into strategic, tactical, and operational offline/online~\cite{Anthony:planning-control,Hans:framework}.

As noted before, the need for and potential of vertically integrated planning approaches is particularly apparent in hospitals. While hospitals are a key player in healthcare systems and account for almost 40\,\% of healthcare spending in OECD countries~\cite{OECD:health-at-a-glance}, they are typically organized as clusters of autonomous departments, and planning is also often functionally dispersed~\cite{Hulshof:PHD}. 
\rev{The care pathways of patients (i.e., the sequences of required activities~\cite{Leeftink_survey}),} however, usually traverse multiple departments~\cite{Hulshof:taxonomy} where different resources are needed for providing effective treatment, which provides a strong motivation for integrated planning of these resources across departments.

\medskip

Consequently, this paper provides the first literature review that focuses specifically on the OR/MS literature related to vertically integrated planning in hospitals. We collect the relevant literature and analyze it with regard to different aspects such as uncertainty modeling and the use of real-life data. Several cross comparisons reveal interesting insights concerning, e.g., relations between the modeling and solution methods used and the practical implementation of the approaches developed. Moreover, we provide a high-level taxonomy for classifying different resource-focused integration approaches and point out gaps in the literature as well as promising directions for future research.

\rev{Table~\ref{tab:comparison-to-other-reviews} provides a comparison of this paper to related review papers. Instead of areas as in the 2010 review by Vanberkel et al.~\cite{vanberkelSurvey} and the 2012 review by Hulshof et al.~\cite{Hulshof:taxonomy}, we consider resources and explicitly integrated planning problems with a focus on hospitals. Moreover, in contrast to the more recent reviews of Marynissen and Demeulemeester~\cite{Marynissen_survey} and Leeftink et al.~\cite{Leeftink_survey} that focus on (multi-)appointment planning, we consider (hospital-based) integrated planning problems in a more general sense, which includes many additional aspects such as capacity dimensioning and staff planning. Compared to~\cite{Marynissen_survey,Leeftink_survey}, this different scope also results in a much larger number of papers included in our review. In summary, the scope of our review is new and provides a useful extension of the already available landscape of review papers.}

\begin{table}[]
    \centering
    \scalebox{.60}{%
    \rev{\begin{tabular}{p{2.3cm}*{5}{p{2.99cm}}}
        \toprule
        & Vanberkel et al.~\cite{vanberkelSurvey} & Hulshof et al.~\cite{Hulshof:taxonomy} & Marynissen and Demeulemeester~\cite{Marynissen_survey} & Leeftink et al.~\cite{Leeftink_survey} & This review\\ \midrule
        \multicolumn{6}{c}{\emph{Structural features and scope}}\\ \midrule
        \# incl.\ papers &  88 & approx.\,400 \hfill (not~expl.~mentioned) & 56 & 63 & >300\\
        Temporal scope & Until 2008 & Until 10 May 2010 & Until the end of 2019	& Until the end of 2016 & Until the end of 2023\\
        Integration focus & Yes & No & Yes & Yes & Yes\\
        Hospital focus & Yes & No & Yes & No & Yes\\
        Search terms provided & No & Yes & Yes & Yes & Yes\\
        Inclusion criteria provided & No & Yes & Yes & No  & Yes\\
        Data bases & Google Scholar & Web of Science (and others, e.g., Scopus) & Web of Science, Scopus & Web of Knowledge, Scopus  & Web of Science, Scopus\\ \midrule 
        \multicolumn{6}{c}{\emph{Content features}}\\ \midrule
        Resources & No (focus on areas) & No (focus on areas) & Yes & No & Yes\\ 
        Methodology & Yes & Yes & Yes & No & Yes\\
        Uncertainty & No & No & No & Yes & Yes\\
        Detailed discussions & Selected Papers  & No & Yes &  Yes & Selected papers\\
        \midrule
        Key differences & Resources not explicitly considered, focus on \emph{areas} & Resources not explicitly considered, focus on \emph{areas}, not focused on hospitals and neither on integrated planning & Focused on appointment planning, does not include other aspects such as capacity dimensioning or staff planning & Focused on appointment planning, does not include other aspects such as capacity dimensioning or staff planning, not focused on hospitals \\ \bottomrule
    \end{tabular}}
    }
    \caption{\rev{Overview of key differences to related reviews. The listed reviews are ordered chronologically from left to right.}}
    \label{tab:comparison-to-other-reviews}
\end{table}

\medskip

The rest of this paper is organized as follows: Section~\ref{sec:search} describes our literature search methodology. The set of relevant papers resulting from the search is then analyzed in Section~\ref{sec:temporal-dev-outlets} regarding the time of publication and publication outlets. Section~\ref{sec:taxonomy} presents our taxonomy for classifying the different vertical integration approaches used in the papers according to three levels of integration. Afterwards, Section~\ref{sec:resource-integration} analyzes which (combinations of) resources are most frequently planned in an integrated fashion. Section~\ref{sec:methods} then focuses on the modeling and solution methods (including methods for uncertainty modeling) that are used for integrated planning, while Section~\ref{sec:data-and-implementation} analyzes the degree of practical implementation achieved by the developed approaches as well as the types of data that are used in the papers. \rev{Subsequently, Section~\ref{sec:state-of-the-art} provides a discussion of recently published, completely integrated planning approaches, and} Section~\ref{sec:hospital+outside} provides an outlook on integrated planning problems that link a hospital to other hospitals and other parts of a healthcare system. \rev{Finally, Section~\ref{sec:conclusion} identifies overarching trends and open research areas, discusses limitations of this review, and concludes our analyses.}

\section{Literature search methodology}\label{sec:search}

To identify relevant literature, an extensive search was performed using the \rev{data\-bases} Web of Science (\url{www.webofscience.com}) \rev{and Scopus (\url{www.scopus.com})}. In order to find papers with an OR focus, the search was performed within journals that are classified as ``Operations Research \& Management Science'' (OR\&MS) according to either their Web of Science Category or their research area (or both) \rev{in Web of Science or as ``Management Science and Operations Research'' according to their All Science Journal Classification Code (ASJC) in Scopus}. Moreover, several relevant journals not classified \rev{accordingly (Computers \& Industrial Engineering, Health Care Management Science, Health Systems, International Journal of Management Science and Engineering Management, Operations Research for Health Care, Simulation)} were identified \rev{based on domain knowledge} and additionally included in the search. To find papers that deal with integrated planning in hospitals, we searched for \rev{original research papers written in English and} published until \rev{until the end of} 2023 for which at least one term from each of the three columns of Table~\ref{tab:search-terms} \rev{must appear} in the title, the abstract, or the author keywords \rev{(i.e., an \emph{and} operator is used to link the hospital, integration, and planning terms categories, while an \emph{or} operator is used for the terms inside each of the three categories)}. Whenever necessary, a wildcard (``\$'' for at most one character or ``$\ast$'' for any group of characters, including no characters) has been used to represent multiple possible endings (e.g., hospital\$ will find ``hospital'' as well ``hospitals'', and integrat$\ast$ will find ``integrate'', ``integrated'', ``integration'' etc.).

\begin{table}[h!]
	\centering
	\begin{tabular}{c|c|c}
		Hospital terms & Integration terms & Planning terms \\
		\hline
		clinic\$                  & collective$\ast$      & allocat$\ast$       \\
		``college\$ of medicine'' & combin$\ast$          & assign$\ast$    \\
		``department\$''          & integra$\ast$         & decid$\ast$   \\
		hospital\$                & join$\ast$            & decision$\ast$      \\
		infirmaries               & multiple              & \rev{design$\ast$}    \\
		infirmary                 & mutual$\ast$          & improv$\ast$     \\
		``medical school\$''      & parallel$\ast$        & transport$\ast$  \\
		surgeo$\ast$              & simultan$\ast$        & manag$\ast$      \\
		surger$\ast$              &                       & maximi$\ast$     \\
		&                    & minimi$\ast$     \\
		&                    & optimi$\ast$     \\
		&                    & plan$\ast$     \\
		&                    & roster$\ast$      \\
        &                    & schedul$\ast$     \\
        &                    & \rev{simulat$\ast$}
	\end{tabular}
    \vspace{2mm}
	\caption{Terms used in the literature search.}
	\label{tab:search-terms}
\end{table}

\medskip

The search returned \rev{2185 results from Web of Science and 2352 from Scopus. After removing duplicates, this resulted in a total of 3469 papers} as search results. \rev{For each of these papers, the title and abstract were then examined by one of the authors in order to exclude papers that are irrelevant. Unclear cases were discussed by all authors.} Here, a paper was excluded if it was clear from the title and abstract that at least one of the following conditions was met: (1) The paper does not focus on hospitals, (2) no integration between multiple resources is considered, or (3) no planning or decision support using any kind of methods from Operations Research and Management Science is considered. Here, following the definition used in~\cite{Hall:handbook}, the term ``resources'' is broadly defined to comprise everything -- from medical and non-medical staff to treatment rooms or patient appointments -- that is required for the provision of healthcare (see Section~\ref{sec:resource-integration} for a classification of different resources considered in the final set of relevant papers). Papers for which it was unclear from the title and abstract whether any of the conditions (1)--(3) are met were \emph{not} excluded here to ensure that no relevant papers are removed from examination at this stage.

\medskip

After the title and abstract screening, \rev{501} potentially relevant papers were left. The full texts of all of these papers were then examined in detail, which resulted in an additional \rev{253} papers that were excluded due to meeting at least one of the above conditions (1)--(3). This resulted in \rev{248} relevant papers that were included in the review.

\medskip

\rev{Additionally, after analyzing the papers included so far, a forward and backward search was performed on a subset of these papers selected according to their number of citations. Since all of the ten most frequently cited papers focus on the operating theater, we decided not to perform the forward and backward search on those papers in order to cover a diverse range of topics. Instead, we selected the single most cited paper from each of five important thematic areas (operating theater, staff, emergency department, outpatients, hospital-wide planning) for the forward and backward search, which returned 986 results. After removing duplicates and papers that had already been found in the initial search, 687 papers were left, which were then examined using the same examination process as described above for the initial search results. This resulted in an additional 71 papers to be added to obtain the final set of 319 relevant papers that were included in the review.} The papers within this final set are listed in a separate bibliography titled ``Search Results'' at the end of the paper \rev{and are classified with respect to all analyzed aspects in the spreadsheet provided as an ancillary file.}

\section{Temporal development and publication outlets}\label{sec:temporal-dev-outlets}
\begin{figure}[ht!]
	\centering
	\includegraphics[clip,width=0.99\linewidth]{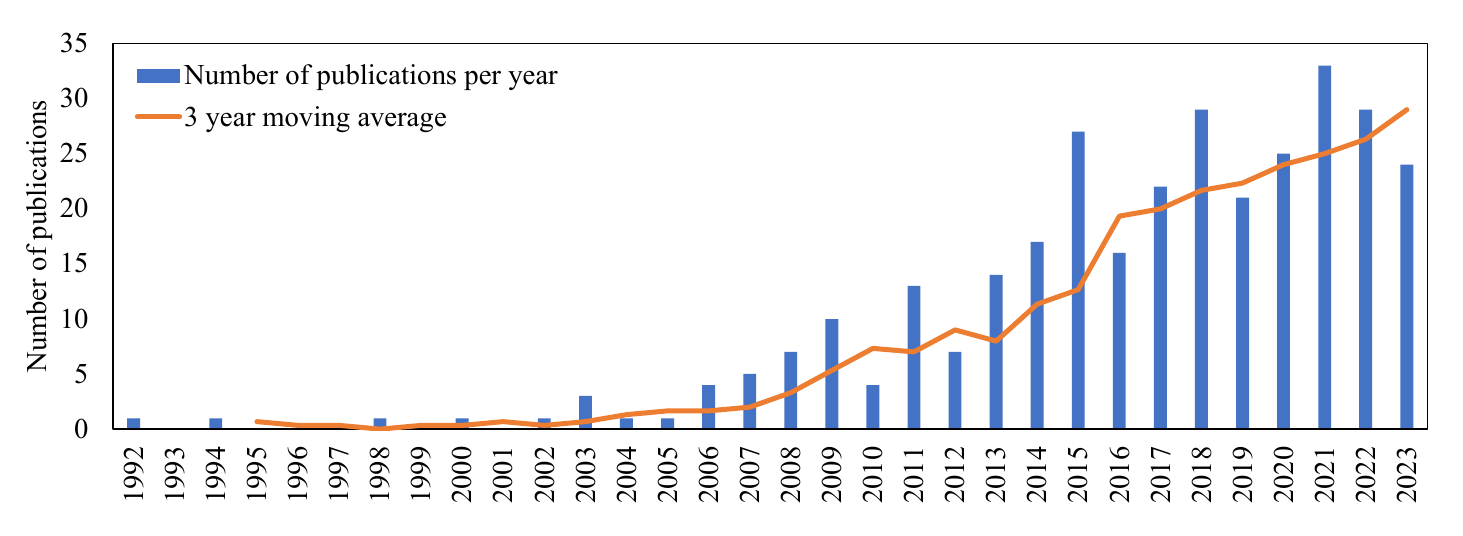}
	\caption{Number of publications over time (per year and 3~year moving average).
    The moving average for year~$t$ considers the years~$t-1$, $t-2$, and~$t-3$.}
	\label{fig:publications-over-time}
\end{figure}

Based on the final set of relevant papers identified, Figure~\ref{fig:publications-over-time} shows the development of the yearly number of publications over time. 
While the first papers on OR/MS in healthcare and hospital contexts have been published in the 1950s~\cite{Hulshof:taxonomy}, the earliest papers on integrated planning in hospitals found in our search stem from the early 1990s, and the yearly numbers of publications show that integrated planning did not receive significant attention in the OR/MS literature until the late 2000s. Since then, the interest in the topic has increased continuously as shown by the 3~year moving average of the number of publications.

\medskip

Concerning publication outlets, most of the relevant papers (\rev{276} of \rev{319}) have been published in journals, while only a small number (\rev{43} of \rev{319}) have been published in conference proceedings. The most frequent publication outlet identified is the European Journal of Operational Research (EJOR) with a total of \rev{42}~published papers, while \rev{the most frequently apprearing journal with a particular focus on OR/MS in healthcare  (Health Care Management Science, HCMS) only has a total of 14~publications, even though it is a well-established journal. The five most frequent outlets also include Computers \& Industrial Engineering~(21), Proceedings of the Winter Simulation Conference~(12), and Computers \& Operations Research~(11).}

\section{Taxonomy of different levels of resource-focused, vertical integration}\label{sec:taxonomy}

In this section, we present a high-level taxonomy for classifying resource-focused, vertical integration approaches, \rev{which is based on our analysis of the approaches} used in the final set of \rev{319} relevant papers that were identified in our literature search (see Section~\ref{sec:search}).
\rev{The taxonomy allows for a concise classification of the large variety of different integrated planning approaches used in the literature according to the degree of resource-focused, vertical integration they achieve. In particular, it allows to compare the achieved degree of vertical integration across approaches that involve different combinations of resources and/or different modeling and solution methods.}

\subsection{Definitions}

In order to classify the approaches for resource-focused, vertical integration used in the set of relevant papers, we categorize them according to the following three \emph{levels of integration}, where a higher level stands for a more closely integrated planning of several resources:

\paragraph{Level 1 (Linkage by constraints / restrictions)} Independent planning of each resource (e.g., staff) that incorporates constraints / restrictions concerning one / multiple other resources (e.g., available beds). These constraints are independent of the concrete solution of the planning problem for the other resource(s).
\paragraph{Level 2 (Sequential planning)} The planning problems for the different resources are solved one after the other in a predefined order (e.g., first staff, then operating room, then beds) and the results of all preceding planning problems (e.g., the staff and operating room plans) are used as input for the planning problem of each resource (e.g., for bed planning). This may or may not include the possibility to return to an earlier planning problem and change this earlier problem's solution using knowledge obtained in later problems (e.g., change the obtained operating room plan since it leads to an infeasible bed planning problem one stage later) -- possibly going back and forth between the problems until the overall process converges (i.e., the solutions for all planning problems satisfy certain quality criteria).
\paragraph{Level 3 (Completely integrated planning)} All resources are planned jointly in one planning problem. Thus, decisions concerning the different resources are made simultaneously and are part of an overall solution of a single problem. 

\medskip

Note that level~1 is conceptually different from levels~2 and~3 in that level~1 approaches do not relate the concrete solutions of the different resource planning problems to each other. By contrast, in level~2 and~3 approaches, the concrete solutions interact either since solutions of preceding planning problems are taken as input when generating the solutions to later planning problems (level~2) or since all solutions are part of an overall solution created in a single joint planning model (level~3). While this means that approaches potentially become more complex with increasing level of integration, the tighter interaction also has the potential to yield better overall solutions.

When considering integrated planning of operating rooms and physicians, for instance, the level~1 approach for master surgery scheduling presented in~\citesearchresults{Cappanera_2014} only takes the availability of surgeons into account via constraints, which ensure that the number of operating room time slots assigned to a given specialty in a given week does not exceed the number of slots that the specialty can cover with the available number of surgeons. Thus, the operating rooms represent the actual planned resource, while physicians (surgeons) are only incorporated via static availability constraints. 
By contrast, the level~2 approach in~\citesearchresults{Day_2012} first assigns blocks of surgery time to surgeons in a first stage. In the second stage, the surgical cases of each surgeon are then assigned a date and a time as consistently as possible with the first stage solution. Finally, in the third stage, the surgical cases are allocated to operating rooms consistently with the solution obtained in the second stage. Thus, the solution for the block scheduling problem of surgeons in the first stage is taken as an input for the operating room planning in the later stages.
Finally, the level~3 approach for operating room scheduling presented in~\citesearchresults{Batun_2011} considers operating room planning and scheduling of surgeons jointly in one model that considers both operating room decisions (e.g., the number of operating rooms to open on a day and the assignment of surgeries to operating rooms) and surgeon decisions (e.g., the start time of each surgeon). Thus, decisions about both resources are taken jointly in one planning model in this case.

\medskip

Note that the distinction between the different planning levels is not completely clear in all cases and there exist papers and approaches that combine several levels -- particularly when considering more than two resources. Overall, completely integrated planning (level~3) and linkage by constraints / restrictions (level~1) are most frequently applied with \rev{198} and \rev{122} papers, respectively, that use approaches of these kinds. Sequential planning (level~2) is far less common with only \rev{11} papers that use planning approaches classified according to this level of integration.\footnote{Note that the single numbers sum up to more than the total number of \rev{319} relevant papers since, as mentioned, some papers present one or several planning approaches with different levels of integration.}

\subsection{Temporal development and relation to hierarchical decision making levels}\label{subsec:temporal-development}

\begin{figure}[ht!]
	\centering
	\includegraphics[width=0.95\linewidth]{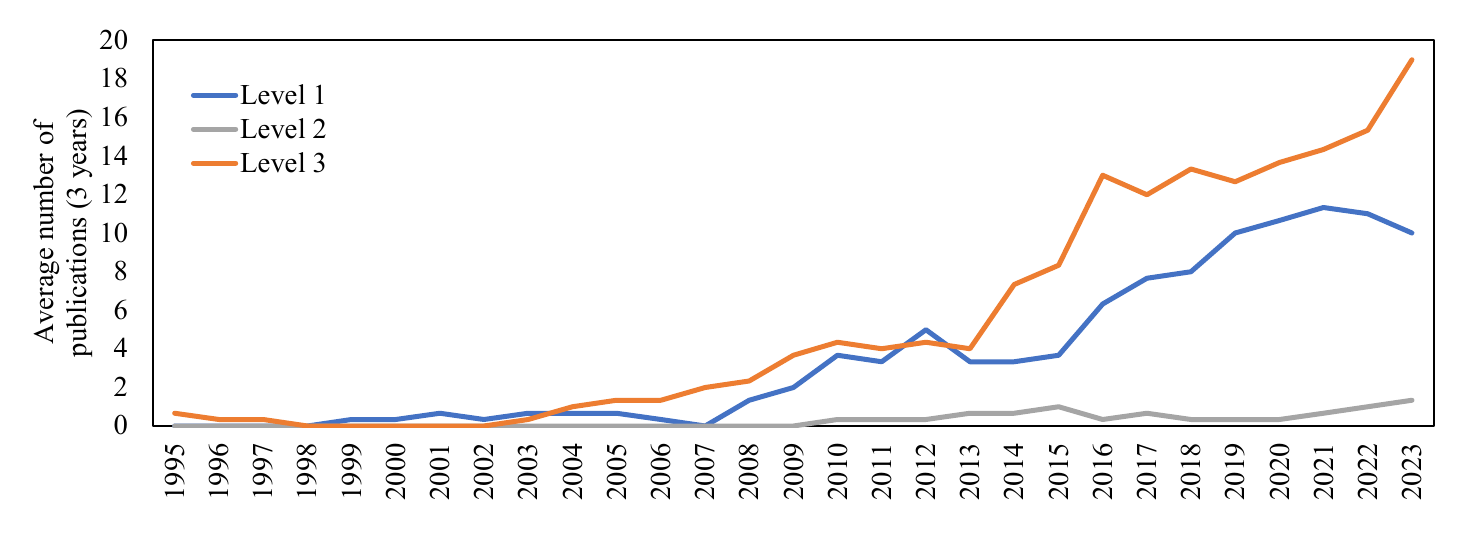}
	\caption{Number of publications as a 3~year moving average distinguished by level of integration. Level~2 is omitted due to the low number of publications.}
	\label{fig:integrated-levels-over-time}
\end{figure}

Figure~\ref{fig:integrated-levels-over-time} shows the temporal development of the 3~year moving averages of the numbers of publications using approaches of the most frequent levels~1 and~3 (level~2 \rev{does not exceed two publications per year}). Interestingly, approaches on both levels of integration started receiving significant attention simultaneously in the late 2000s. They have first been similarly common with \rev{26} papers using level~1 approaches and \rev{31} papers using level~3 approaches before 2012. Among the papers published since 2012, however, only \rev{96} use level~1 approaches, while \rev{167} use level~3 approaches. This means that completely integrated approaches that plan several resources jointly in one model have become more popular compared to approaches that plan each resource independently while only incorporating constraints or restrictions concerning other resources. This trend towards completely integrated planning approaches could be explained by both a rising interest in deeper integration between planning problems of different resources but also by increasingly powerful computers and solvers, which make completely integrated models solvable in more reasonable times than before.

Next, we investigate how the different levels of integration relate to the well-known hierarchical levels of strategic, tactical, operational offline, and operational online decision making~\cite{Anthony:planning-control,Hans:framework}. 
Table~\ref{tab:planning-horizon-vs-integration} shows the numbers of publications for each of the hierarchical levels in total as well as distinguished by level of integration.\footnote{Again note that the single numbers sum up to more than the total number of \rev{319} relevant papers since some papers present several planning approaches targeting different hierarchical decision making levels and / or different levels of integration.}
The table shows that, for both level~1 and level~3 integration, most papers target operational problems, the vast majority of which are operational offline. Tactical integrated planning problems are studied less frequently on both levels of integration, and even fewer publications consider a strategic planning horizon. \rev{The few existing level~2 approaches are mostly focussed on tactical and operational offline problems, while strategic and operational online problems receive only very little attention.}

\begin{table}\centering%
	\begin{tabular}{lcccccc}
		\hline
		& Strategic & Tactical & \multicolumn{3}{c}{Operational}                        & \rev{Overall} \\
		&           &          & \emph{total} & \emph{offline} & \emph{online} &       \\ \hline 
		Total   & \rev{31}     & \rev{108}      & \rev{221}       & \rev{190}      & \rev{31}    & \rev{360}   \\ \hline 
		Level 1 & \rev{11}     &  \rev{35}      &  \rev{92}       &  \rev{77}      & \rev{15}    & \rev{138}    \\
		\rev{Level 2} & \rev{1}      &  \rev{7}       &   \rev{7}       &   \rev{6}      &  \rev{1}    &  \rev{15}    \\
		Level 3 & \rev{25}     &  \rev{71}      & \rev{126}       & \rev{111}      & \rev{15}    & \rev{222} \\ \hline
	\end{tabular}
    \vspace{2mm}
	\caption{Hierarchical decision making level versus level of integration.}
\label{tab:planning-horizon-vs-integration}
\end{table}

\section{Resources considered in integrated planning approaches}\label{sec:resource-integration}
\begin{figure}
	\centering
 \includegraphics[width=0.75\linewidth]{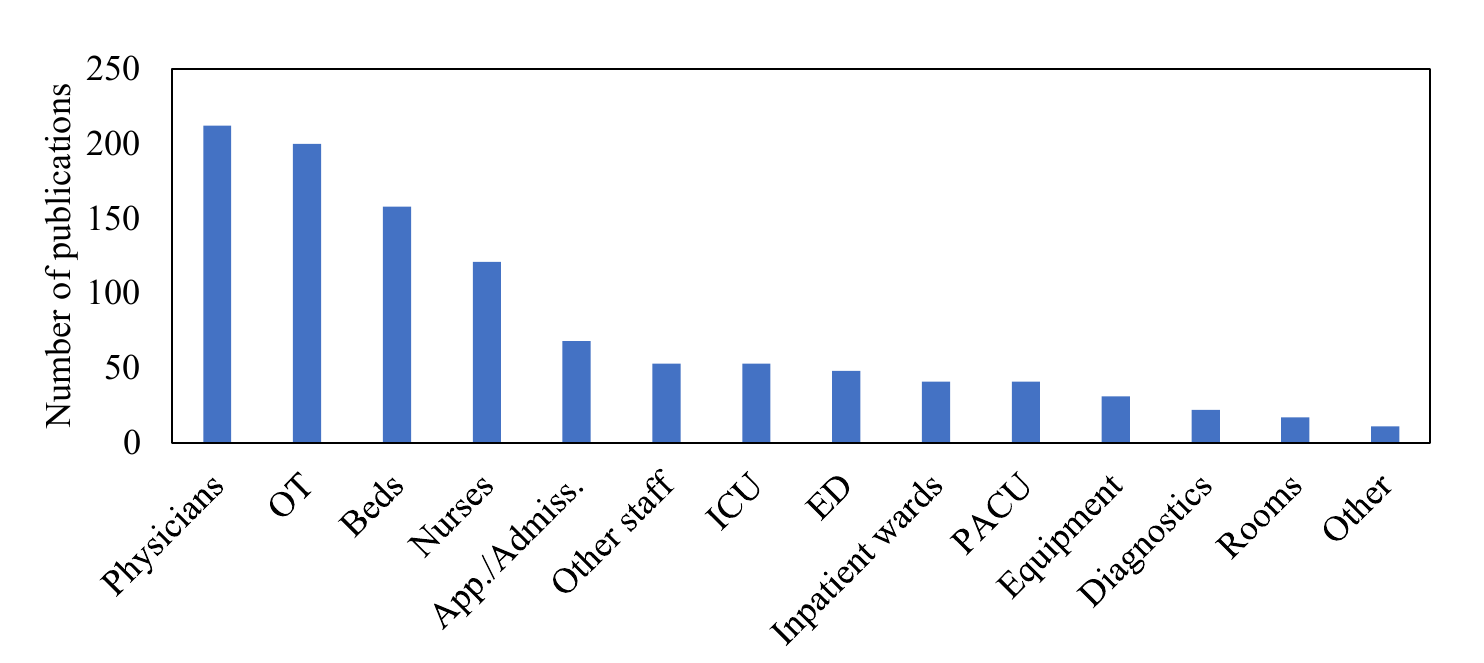}
	\caption{Absolute frequencies of considered resources overall. \emph{App./Admiss.} and \emph{Rooms} are used to abbreviate \emph{patient appointments~/~admissions} and \emph{examination / treatment rooms}, respectively\rev{. \emph{Diagnostics} is used as an umbrella term for radiology, imaging services (CT, MRI, X-Ray), and laboratory}, while \emph{Other} summarizes all further resources that occurred too infrequently to warrant a separate listing \rev{(clinical services, elevators, logistics, outpatient clinics, physiotherapy, and sterilization services)}.}
	\label{fig:integrated-resources1}
\end{figure}

Having demonstrated the growing interest in integrated planning problems over the last two decades, we now analyze which resources have been at the center of attention. Initially, Figure~\ref{fig:integrated-resources1} shows the absolute frequencies of hospital resources / areas considered in integrated planning approaches. The figure shows that the vast majority of publications deal with \rev{medical staff (physicians and nurses), the operating room / operating theater (OT)\footnote{We use OT as an abbreviation for consistency reasons since OR is used as an abbreviation for Operations Research.}}, or beds. Other frequently considered resources include patient appointments~/~admissions, \rev{other staff (e.g., technicians}), intensive care unit (ICU), \rev{emergency department (ED), inpatient wards,} and post-anesthesia care unit (PACU). \rev{Here, the resources ICU, ED, inpatient wards, and PACU refer to general capacity and/or costs of these areas, which are often not defined precisely in the relevant papers and can result from different considerations such as throughput limitations or combinations of bed, staff, and medical equipment availability.} It is notable that \rev{resources such as clinical services (which includes hospital pharmacy), logistics, or sterilization that are not directly involved in patient-related tasks (surgery, caring on a ward, etc.) have only received limited attention. In other words, resources and problems that are (further) away from the patient are rarely studied in integrated planning approaches so far. }

\begin{figure}
	\includegraphics[width=0.99\linewidth]{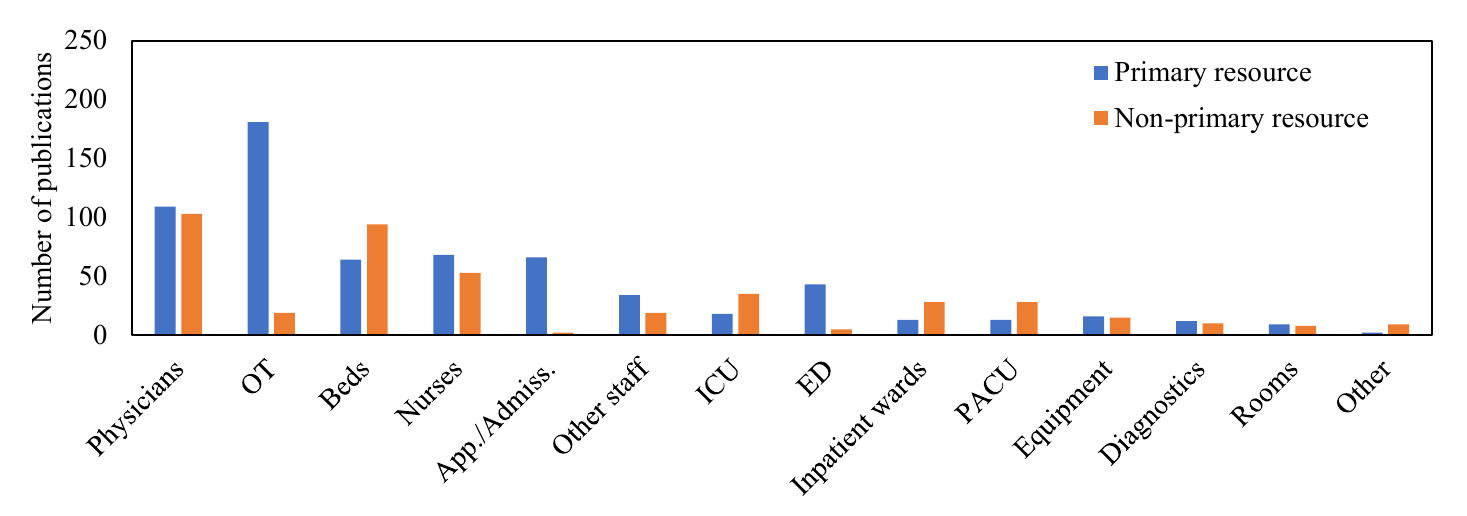}
	\caption{Absolute frequencies of considered resources distinguished by their importance (primary, non-primary).}
	\label{fig:integrated-resources2}
\end{figure}
The analysis in Figure~\ref{fig:integrated-resources2} distinguishes between primary and non-primary resources, i.e., resources / areas that are at the center of planning and ones that are supplementary. Note that, if several resources are integrated using a level~3 integration approach (see previous section), all of these resources are usually considered as primary or leading resources, but additional non-primary resources can also be included (e.g., via constraints using a level~1 integration approach). OT, patient appointments / admissions, and ED are regularly considered as primary resources, while only a small proportion of the papers containing these resources considers them as a supplement to other resources. In contrast, bed-related resources (including wards, PACU, and ICU) are more frequently considered as supplementary than primary. With \rev{physicians,} nurses, medical equipment, diagnostics, and examination / treatment rooms, the results are balanced. Interestingly, while medical staff (physicians and nurses) \rev{are} \rev{almost balanced,} other staff (including porters and technical staff) \rev{are} mostly planned as a primary resource.

\begin{figure}
	\centering
	\includegraphics[width=0.99\textwidth]{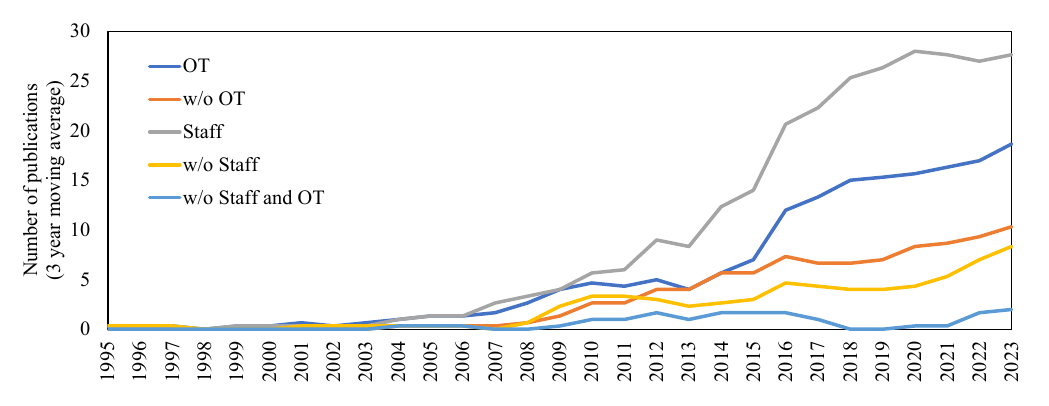}
	\caption{3 year moving averages of the numbers of publications that do / do not consider OT or medical staff.}
	\label{fig:trends-over-time}
\end{figure}

\rev{In line with operating theaters being the most-studied area of the hospital in the OR/MS literature in general~\cite{vanriet2015:trade-offs-or,guerriero2011:survey-or,Cardoen:operating-room-survey}, the joint consideration of Figures~\ref{fig:integrated-resources1} and~\ref{fig:integrated-resources2} shows that the OT is both one of the two most frequently considered resources in integrated planning approaches overall as well as the resource that is most frequently considered as primary within these integrated approaches. When combining physicians and nurses under the umbrella term \emph{medical staff}, however, the temporal development illustrated in Figure~\ref{fig:trends-over-time} shows an interesting trend. The different lines visualize the 3~year moving averages of the yearly numbers of publications that do / do not consider the OT or medical staff as one of the integrated resources. Despite the large number of OT-focused publications in the OR/MS domain overall, Figure~\ref{fig:trends-over-time} suggests that, since about 2010, the number of publications considering medical staff within integrated planning approaches constantly exceeds the number of publications considering the OT. The OT, however, is still the most frequently appearing primary resource even when counting physicians and nurses jointly. This suggests that the OT is still the most common center of attention also in integrated planning approaches, but medical staff \rev{have} been considered more frequently as part of integrated planning approaches overall for more than a decade. This is also supported by the observation that the number of papers that do \emph{not} include medical staff is lower compared to the number of papers that do \emph{not} include the OT -- both in total and relative to the number of papers that \emph{do} include the respective resource. A possible explanation could stem from increasing shortages of medical staff, which might motivate to at least include staff as a supplementary resource in integrated planning problems.}

\rev{When considering the number of publications that include neither the OT nor medical staff, it can be observed that the average number of such publications has stayed extremely low overall, even though the average number of publications per year has increased tremendously within the last 15 years (see Figure~\ref{fig:integrated-levels-over-time} in Section~\ref{subsec:temporal-development}).}

\subsection{Considered resource combinations}
\begin{figure}
	\centering
	\includegraphics[width=0.7\linewidth]{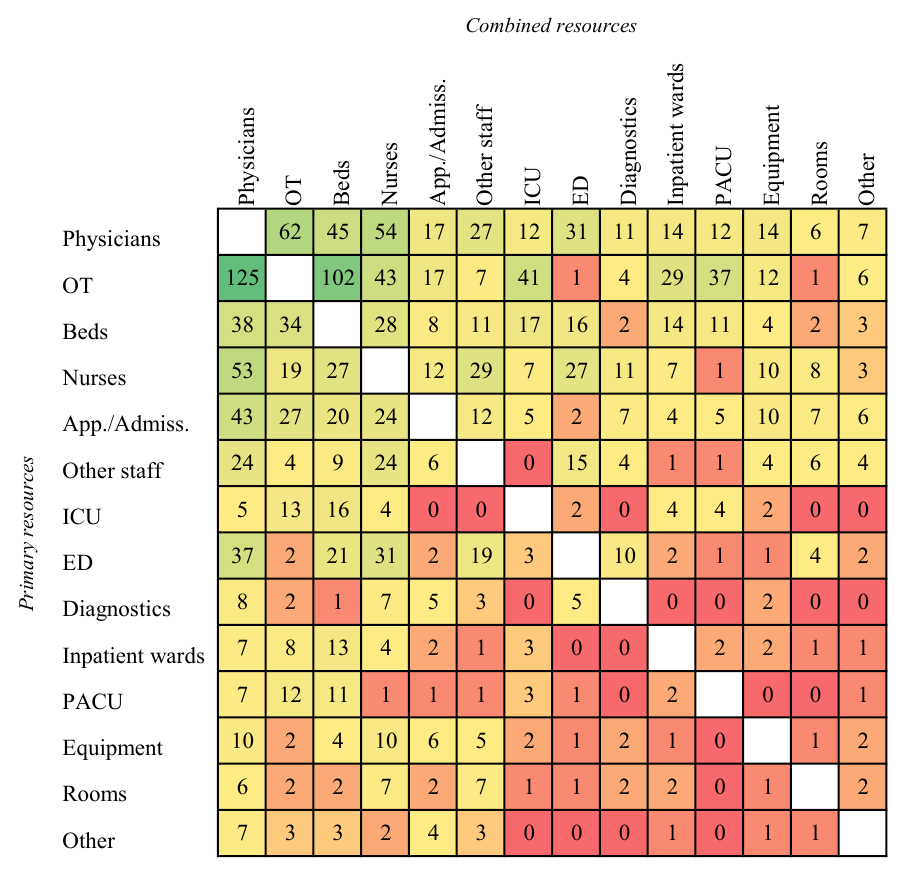}
	\caption{Heat map indicating absolute frequencies of resource / area combinations. Rows correspond to primary resources and columns to combined resources that can be either primary or non-primary. For example, the number \rev{27} in the cell within the row \emph{Nurses} and column \emph{Beds} indicates the number of publications in which nurses are the primary resource and beds are considered as either primary or non-primary.
	}
	\label{fig:heatmap-basic}
\end{figure}

We now look at specific combinations of resources that are considered and further detail the analysis regarding primary and non-primary resources. We use the same categories that have previously been introduced in Figure~\ref{fig:integrated-resources1}. Figure~\ref{fig:heatmap-basic} displays the absolute frequencies of individual combinations of resources in a heat map linking \emph{primary resources} in rows to \emph{combined resources} (which can be either primary or non-primary) in columns. The number in each cell indicates the number of publications in which a link between the two corresponding resources is found. The background of each cell is color-coded ranging from dark green (highest absolute frequency) to red (absolute frequency zero). Combinations of a resource with itself are excluded for obvious reasons. 

The heat map in Figure~\ref{fig:heatmap-basic} reveals that \emph{OT (primary) \& physicians (combined)} and \emph{OT (primary) \& beds (combined)} are by far the most common combinations with absolute frequencies of~\rev{125} and~\rev{102}, respectively. This is in line with our previous observation that the OT is mostly considered as a primary resource. Additionally, the fact that \rev{beds} are so frequently combined with the primary resource OT provides a possible explanation why \rev{this resource is} not the primary resource in the majority of papers considering \rev{it}. A more balanced distribution of which of the two considered resources is primary can be observed, e.g., for the combination \rev{of} nurses \rev{and} physicians, where each of the two is considered primary in \rev{53 and 54}~papers\rev{, respectively}. 

\subsection{Resource combinations and levels of integration}
\begin{figure}
	\centering
	\includegraphics[width=0.75\textwidth]{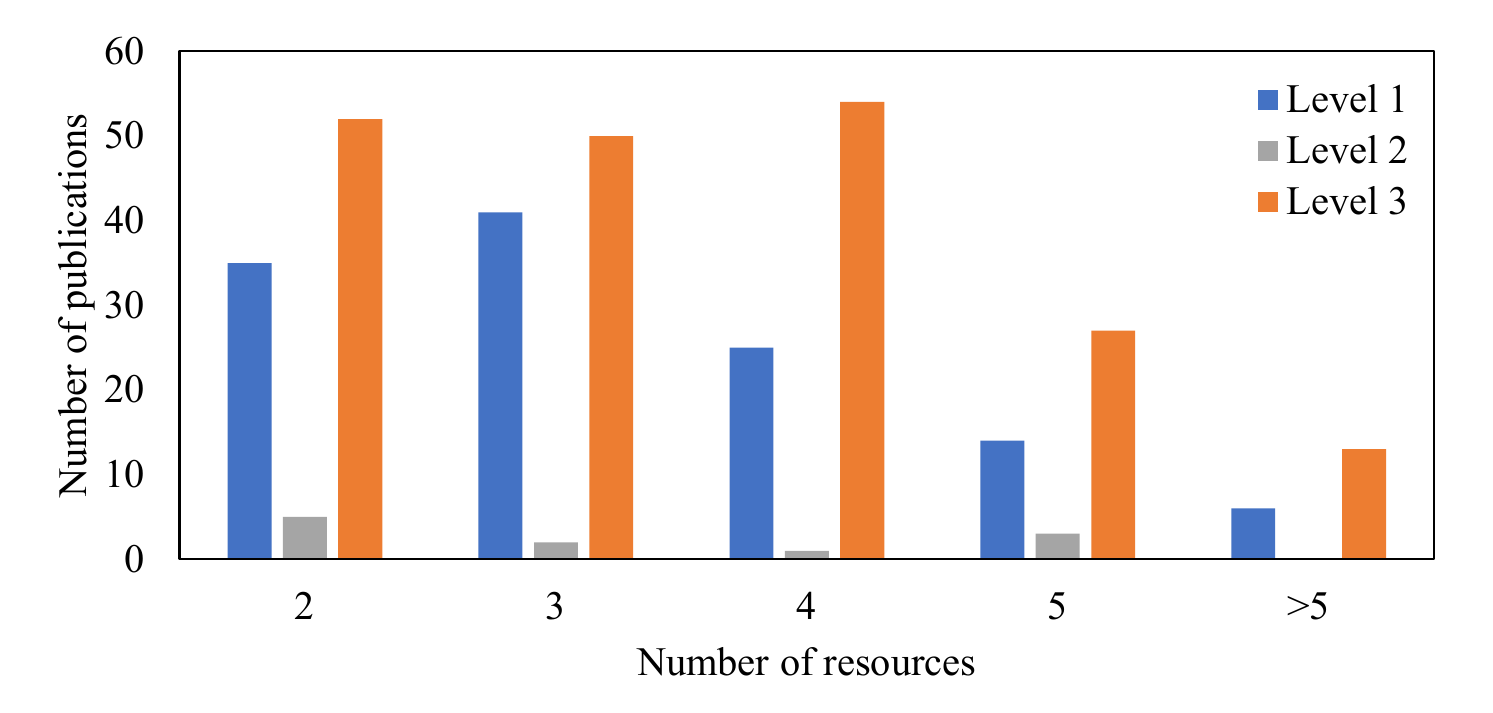}
	\caption{Absolute frequencies of different numbers of integrated resources differentiated by level of integration.}
	\label{fig:number-of-resources}
\end{figure}

\begin{figure}
	\centering
	\includegraphics[width=1\linewidth]{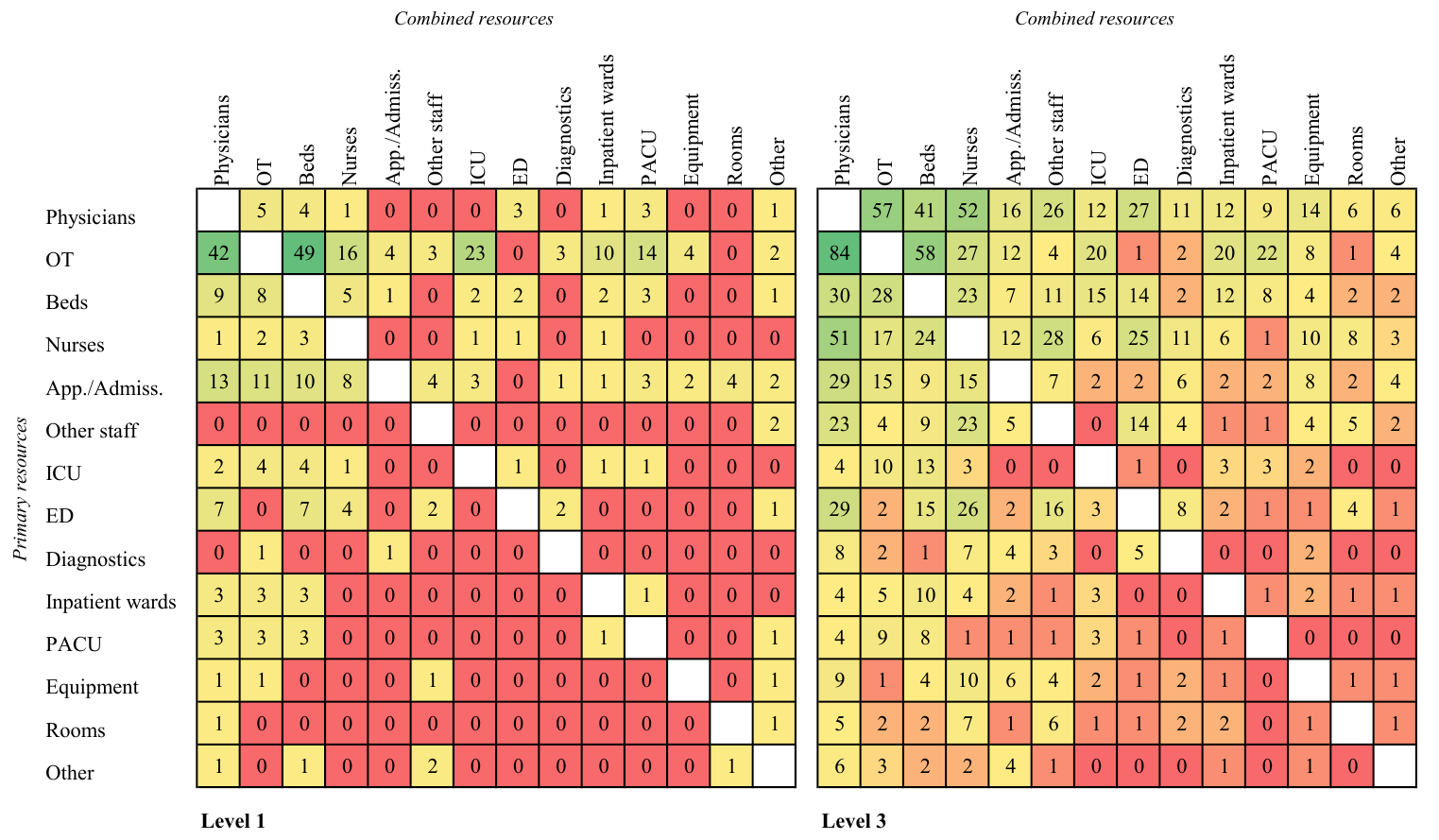}
	\caption{Heat maps indicating absolute frequencies of resource / area combinations for level~1 integration (left) and level~3 integration (right).}
\label{fig:heatmap-level1and3}
\end{figure}

We now further distinguish the considered resource combinations with regard to the level of integration. Figure~\ref{fig:number-of-resources} shows the absolute frequencies of different numbers of integrated resources differentiated by level of integration (\rev{despite its low abosolute frequency, we include} level~2). 
While \rev{up to four} integrated resources are most common in both level~1 and level~3 approaches, \rev{five} or more resources are integrated in \rev{only approximately 6\% and 13\% of the cases, respectively. For level~3, all numbers of integrated resources up to four have almost equal frequencies, while two and three integrated resources appear more frequently for level~1. The few existing level~2 integration approaches consider at most five resources, without showing a specific trend.}
For the sake of a cross comparison of specific resource combinations and the considered level of integration, Figure~\ref{fig:heatmap-level1and3} depicts two heat maps analogous to the one from Figure~\ref{fig:heatmap-basic} -- one for level~1 approaches and one for level~3 approaches.
Recall that the number of publications for level~3 is larger than for level~1 (\rev{198} vs.~\rev{122} publications). Focusing on level~1 integration (Figure~\ref{fig:heatmap-level1and3}, left), there is less variety in considered resource combinations (indicated by many zeros) compared to level~3 integration. For level~1 integration, the OT is by far the most common primary resource, which is frequently combined with beds / ICU capacity or staff (nurses and physicians). In contrast, the right-hand heat map in Figure~\ref{fig:heatmap-level1and3} indicates a much larger variety of resource combinations for level~3 integration, with staff-focused papers in particular appearing more frequently. Overall, the cross comparison of specific resource combinations and the considered level of integration reveals that the previously-observed dominance of the OT as a primary resource is mainly due to it being so frequently considered as the main resource in level~1 integration approaches, which often address OT planning with constraints concerning the availability of beds / ICU capacity and / or staff. While the OT is still often considered as as primary resource in level~3 approaches, there is also a large number of level~3 approaches that consider staff -- particularly physicians -- as a primary resource. In a relevant number of level~3 approaches, we can even observe that physicians or nurses are considered as a primary resource even when combined with the OT. Since other (non-medical) staff, which \rev{are} ignored almost completely in level~1 approaches, \rev{are} also considered much more frequently in level~3 approaches, this could indicate a shift from OT-focused integrated planning in purely constraint-related integrated planning approaches to a more staff-focused planning in approaches that perform on a completely integrated planning of several resources.

\section{Modeling and solution methods}\label{sec:methods}
\begin{figure}
	\centering
	\includegraphics[width=0.9\textwidth]{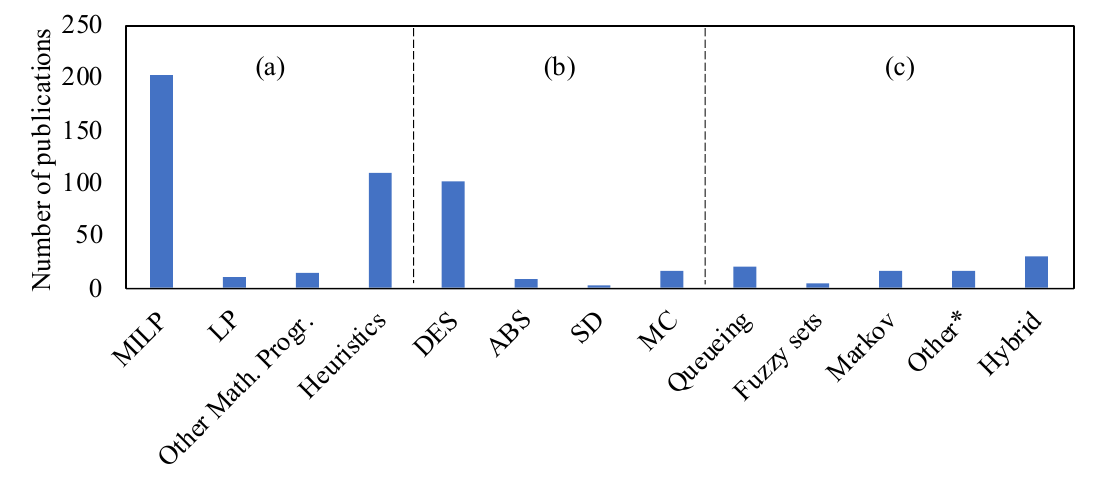}
	\caption{Absolute frequencies of methods by categories: (a)~optimization approaches, (b)~simulation approaches, and (c)~other methods. The following abbreviations are used: \emph{MILP} (Mixed-Integer Linear Programming), \emph{LP} (Linear Programming), \emph{DES} (Discrete Event Simulation), \emph{ABS} (Agent-Based Simulation), \emph{SD} (System Dynamics), \emph{MC} (Monte Carlo Simulation). \rev{The term \emph{Hybrid} in~(c) refers to hybrid methods resulting as combinations of several different methods, while} the term \emph{Other*} summarizes all further methods that occur too infrequently to warrant a separate listing.}
	\label{fig:methods-overview}
\end{figure}

When integrating healthcare planning problems, which are often already complex individually, they potentially grow in size and, thus, might become even more difficult to solve. Therefore, choosing adequate modeling and solution methods is a highly relevant aspect.  Figure~\ref{fig:methods-overview} highlights the various approaches applied in the publications, aggregated into (a)~optimization (\rev{238}), (b)~simulation (\rev{125}), and (c)~other methods, e.g., queuing theory or machine learning (\rev{76}). Among optimization-focused publications, \rev{202} of the publications use mixed-integer linear programming (MILP), while linear programming (LP) or \rev{further} mathematical programming techniques (e.g., quadratic programming) are rarely used (\rev{29} in total). Among the simulation paradigms, discrete event simulation~(DES) is most popular with \rev{102}~publications, while the \rev{altenative} paradigms such as agent-based simulation~(ABS), system dynamics~(SD), or Monte Carlo simulation~(MC) play only a minor role (\rev{36} papers in total). Lastly, other methods summarized in~(c) predominantly focus on queuing and Markov models, while different concepts such as fuzzy sets or scattered use of methods such as machine learning hardly occur. Here, \rev{34} papers using hybrid approaches indicate tailored modeling/solution concepts such as simulation-optimization or a combination of queuing/simulation and Markov/simulation. 
In Table~\ref{tab:methods-combined}, we visualize whether and to what extend two or more methods have been applied for solving an integrated planning problem. For optimization problems, it is common to apply only a single method when solving a problem. For simulation, however, it is more common to combine simulation with either optimization~(\rev{51}), other methods~(\rev{20}), or both~(\rev{14}) than to use a simulation-only approach~(\rev{40}). However, the use of multiple methods is typically sequential. \rev{When combining simulation and optimization, for example, common approaches include using simulation to generate scenarios as input for an optimization model (see, for instance,~\citesearchresults{Min_2010,Wang_2018}) or evaluating the solution obtained from of an optimization model using simulation (see, for instance,~\citesearchresults{Deklerck_2021,Bovim_2022,Daldoul_2022,Dehghanimohammadabadi_2023,Rachuba_2022}).}
\begin{table}[]
	\centering
	\begin{tabular}{l|ccc|c}
		& Optimization &  Simulation & Other methods & All  \\ \hline
		Optimization   & \rev{152} & \rev{51} & \rev{21} & \\
		Simulation     & \rev{51}  & \rev{40} & \rev{20} & \\
		Other methods  & \rev{21}  & \rev{20} & \rev{20} &\\ \hline 
		All & & & & \rev{14}
	\end{tabular}
    \vspace{2mm}
	\caption{Absolute frequencies of methods applied and possible combinations. Values on the diagonal indicate that only one kind of method is used, e.g., purely simulation. ``All'' indicates that methods of each kind are used in a publication.}
	\label{tab:methods-combined}
\end{table}

\begin{figure}
	\centering
	\includegraphics[width=0.99\textwidth]{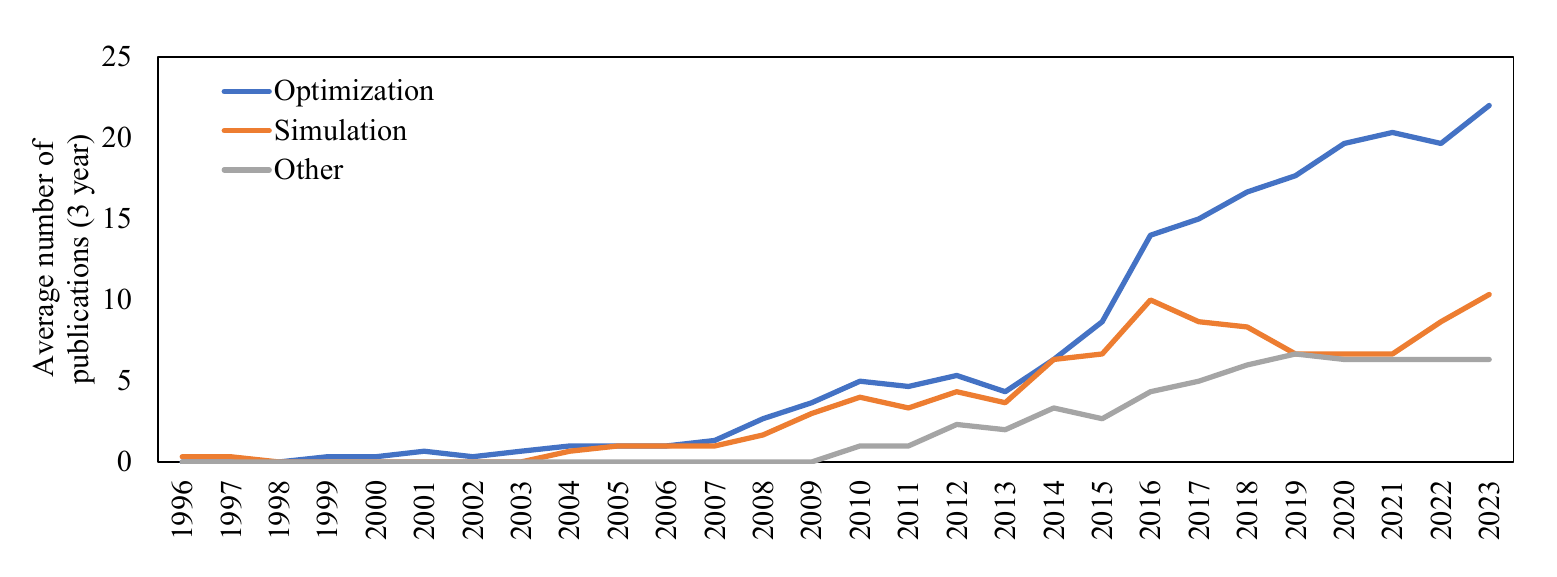}
	\caption{High-level overview of methods over time (3~year moving averages).}
	\label{fig:methods-over-time-1}
\end{figure}

\textcolor{black}{Figure~\ref{fig:methods-over-time-1} visualizes the evolution of the average number of publications by method categories. It highlights a steady increase of optimization-based studies since approximately 2010. Only twice we identify a decline, namely in 2013 and between \rev{2021} and 2022, but only in small volume. Since 2022, an increase in the number of publications is observed again. Up to \rev{2016}, we observe a similar trend for simulation-focused publications, which did not increase (and even \rev{decreased}) towards 2021. Since 2021, however, we identify a strong increase in the volume of publications. 
Lastly, other approaches start to appear from the late 2000s, with an increase to approximately \rev{6}~papers on average in 2018. This level is maintained to the present day.}

\subsection{Cross comparisons with regard to modeling and solution methods}
\begin{figure}
	\centering
	\includegraphics[width=0.99\textwidth]{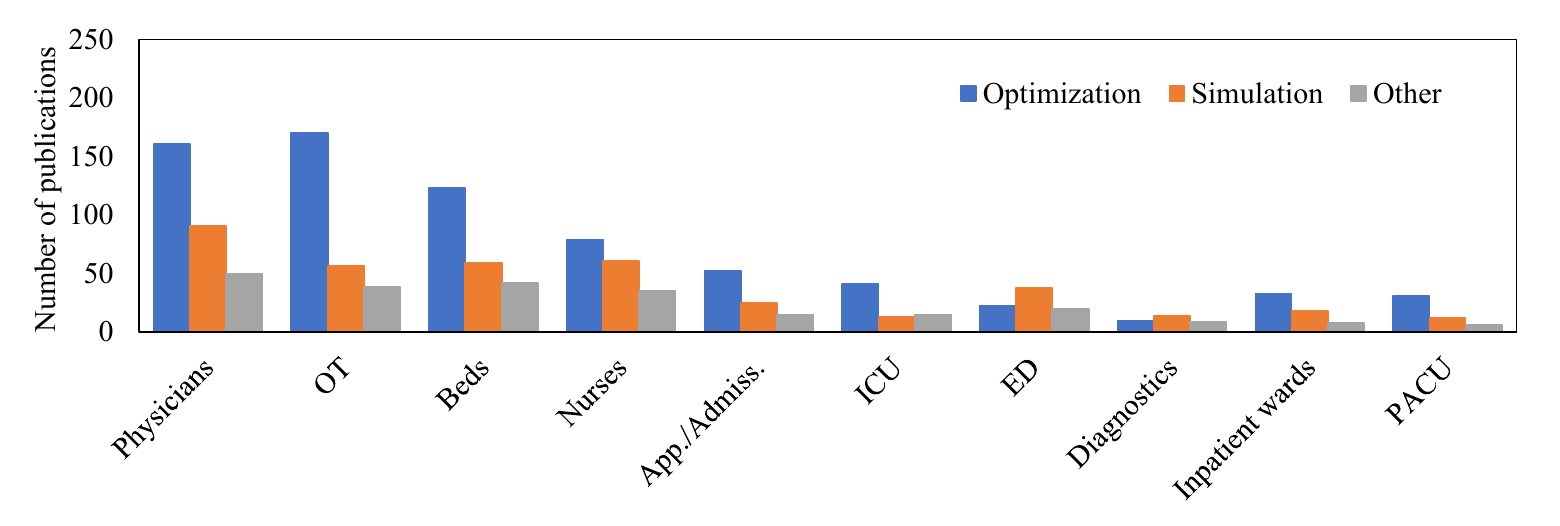}
	\caption{High-level overview of methods by most commonly found resources.}
	\label{fig:methods-by-resources-0}
\end{figure}

\begin{figure}
	\centering
	\includegraphics[width=0.99\textwidth]{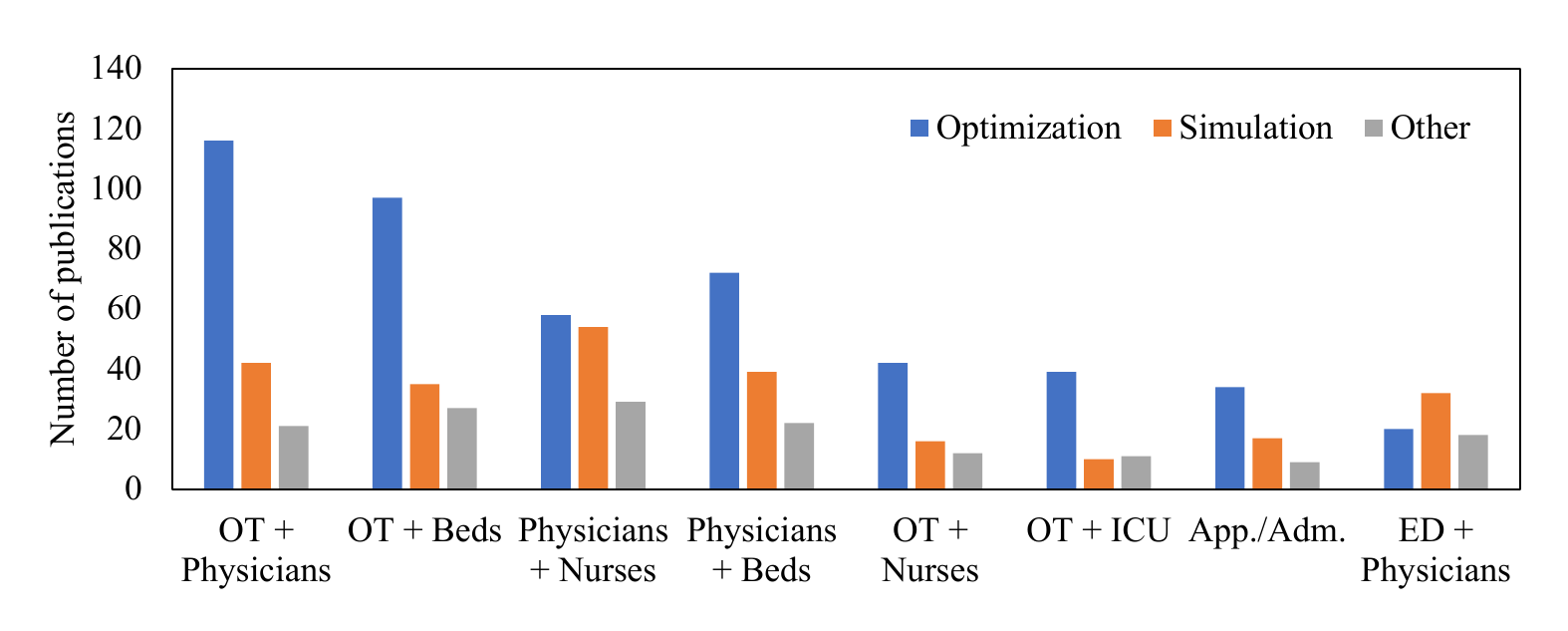}
	\caption{High-level overview of methods by most commonly found \emph{combinations of} resources.}
	\label{fig:methods-by-resources-1}
\end{figure}

In the following, we analyze the applied modeling and solution methods depending on (1)~planned resources, (2)~level of integration, and (3)~hierarchical decision making level. Similar to Table~\ref{tab:methods-combined}, we consider the aggregated method categories optimization, simulation, and other methods. 
Figure~\ref{fig:methods-by-resources-0} shows absolute frequencies of planned resources distinguished by the three categories of methods \rev{(the three smallest categories (equipment, rooms, and other) are omitted due to small numbers)}. \rev{We observe that} optimization is \rev{usually} preferred over simulation or other approaches, which is in line with a larger share of optimization studies (see also Figure~\ref{fig:methods-overview}). For \rev{physicians, OT, and also beds}, the share of optimization-focused papers is disproportionately larger compared to other resources. Especially for the OT, the relative frequency of approaches using simulation or other methods is relatively small. While there are mostly fewer simulation papers, the majority of ED-related papers do use simulation.
These findings also show when comparing the most common combinations of two resources (see Figure~\ref{fig:methods-by-resources-1}). Together with the previous results, we can conclude that OT-related publications predominantly use optimization, even when staff-related resources are considered as well, e.g., OT and physicians. What is interesting to note is the fact that, for purely staff-related combinations such as \rev{physicians and nurses}, it is much more common to use simulation and also other approaches. In general, when no staff \rev{are} involved, optimization is chosen more frequently. For the combination of ED and physicians, the share of optimization papers is \rev{notably smaller than the share of simulation papers, and only slightly larger than the share of papers using other methods.} It is also worth noting that a combination of OT and ICU leads to the smallest share of simulation studies \rev{for this selection}. 

\begin{figure}[h!]
	\centering
	\includegraphics[width=0.95\textwidth]{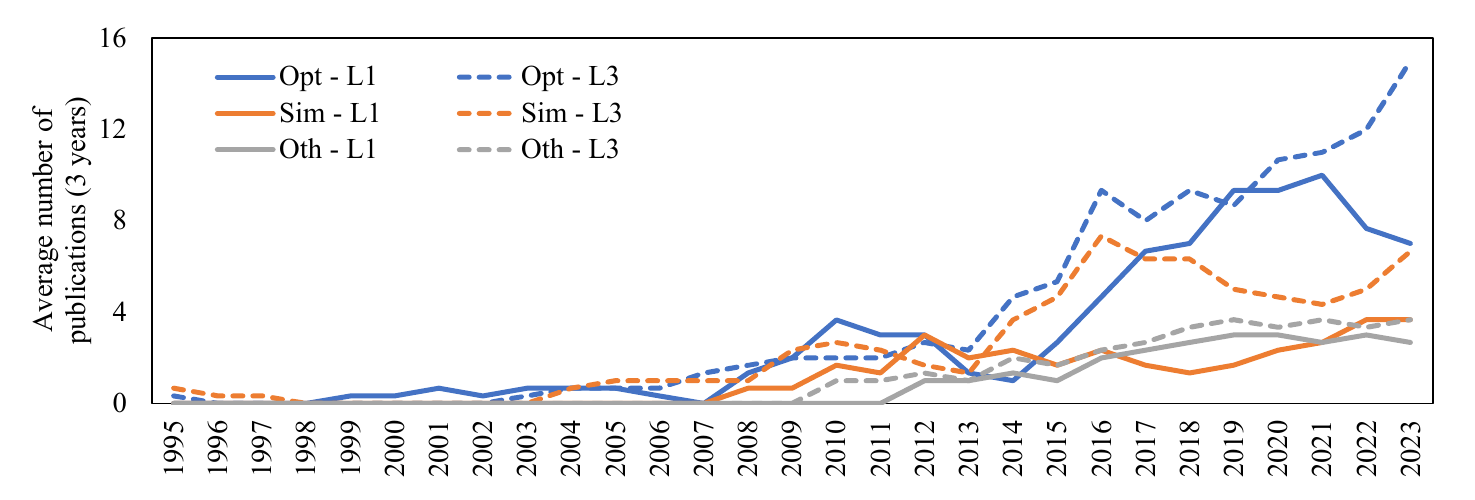}
	\caption{High-level overview of methods over time (3 year moving averages) distinguished by the level of integration.}
	\label{fig:methods-over-time-2}
\end{figure}

While there has been an almost steady increase in optimization papers over the last two decades (see Figure~\ref{fig:methods-over-time-1}), simulation and other approaches do not \rev{consistently} exceed approximately 5 papers in a 3 year average.  Figure~\ref{fig:methods-over-time-2} now investigates this further and depicts the development of method categories over time distinguished by the level of integration that the publications consider.  What is interesting in this case is the fact that, in recent years, optimization-based studies see an increase in level~3 studies (blue line, dashed) while level~1 studies appear less frequently. For the two other method categories (orange and gray lines), however, we see a more similar development when comparing level~1 and level~3 approaches. For optimization, there has been a clear peak in the number of level~1 approaches between 2019 and \rev{2021} followed by a strong decrease. Level~3 optimization approaches, however, have seen an almost steady increase, and a particularly strong increase since \rev{2019}. This could potentially be explained by computational advancements that now make it possible to solve the typically more complex level~3 optimization models in more reasonable time than before.

Table~\ref{tab:methods-overview} shows the numbers of publications using each method category distinguished by hierarchical decision making level. Note that papers using several methods from different categories are counted multiple times. Concerning hierarchical decision making levels, optimization approaches dominate on \rev{all levels, but most notably on the operational level, where they are far more frequent overall than simulation and other approaches combined.}
\begin{table}
	\centering
	\begin{tabular}{l|ccc}
		Methods &  Optimization &  Simulation & Other methods  \\ \hline
		Total         & \rev{238}    & \rev{125}   & \rev{76}\\ \hline 
		Strategic     & \rev{22}     & \rev{16}    & \rev{8} \\
		Tactical      & \rev{75}     & \rev{50}    & \rev{30} \\
		Operational   & \rev{172}    & \rev{79}    & \rev{44} \\
		~~\emph{offline/online}%
                      & \rev{150/22} & \rev{69/10} & \rev{34/10}\\ \hline
	\end{tabular}
    \vspace{2mm}
	\caption{ Absolute frequencies of method categories used overall and distinguished by hierarchical decision making level.}
\label{tab:methods-overview}
\end{table}

Among the papers that use optimization approaches, \rev{173} focus on a singleobjective approach while only \rev{61} consider a dedicated multiobjective approach. For level~3 problems, multiobjective approaches are slightly more common (\rev{37 out} of \rev{142} at this level) compared to level~1 problems (\rev{23 out} of \rev{97} publications). \rev{With respect to hierarchical decision making levels, the share of multi- versus singleobjective approaches is lower on the strategic level (\rev{4} versus \rev{18}) compared to the tactical (\rev{21} versus \rev{52}) and the operational level (\rev{43} versus \rev{116}).} 
What is potentially most interesting is the fact that, \rev{among the optimization studies using multiobjective appraoches, only five~\citesearchresults{Testi_2007,Park_2021,Bentayeb_2023,Calegari_2020,Heider_2022_a} report} practical applications of the developed results or methods (see Section~\ref{sec:data-and-implementation} for more analyzes concerning practical implementation).

\subsection{Uncertainty modeling}
Most real-world planning problems in hospitals suffer from uncertainty, i.e., from incomplete information regarding some of the problem parameters or input data. Therefore, dealing with uncertainty is an important aspect of these problems. We now analyze the identified literature concerning the approaches used for dealing with uncertainty. Common ways to model uncertainty are stochastic models, robust models, and online \rev{optimization (competitive analysis)}~\cite{Bakker_2020}. While the classifications \emph{robust} and \emph{online} apply only to optimization approaches, the classification \emph{stochastic} can be considered for optimization, simulation, or other approaches, e.g., Markov or queuing models. In contrast to all three of these classifications, deterministic models assume all problem parameters and input data to be completely known without any uncertainty at the time the problem is solved.

Table~\ref{tab:uncertainty-and-integration} shows the overall distribution of publications differentiated by uncertainty modeling approach and level of integration. Note that a paper is counted twice if, e.g., both a robust \emph{and} a stochastic model are presented in the corresponding publication. Values in bold font indicate the total numbers of papers (e.g., there are \rev{89}~publications in total that use deterministic planning approaches for level~3 integration) and values in parentheses indicate the numbers of papers with/without an optimization model (e.g., of the previously mentioned \rev{89}~papers that use deterministic level~3 planning approaches, \rev{83} use an optimization model, while \rev{6} do not).

Overall, \rev{the majority} of papers considers uncertainty in some of the input data (\rev{193} of \rev{319}). Interestingly, the share of papers considering uncertainty is \rev{slightly} larger among papers presenting level~1 integration approaches (\rev{76} of \rev{122}) than among papers presenting level~3 approaches (\rev{119} of \rev{198}), even though \rev{level~3} approaches are more recent on average (see Section~\ref{subsec:temporal-development}). A possible reason for this could be that completely integrated models are potentially harder to solve than models that only incorporate further resources using constraints, which could mean that considering uncertainty as well might not always be tractable in completely integrated models.
\begin{table}
	\centering
	\small
	\begin{tabular}{l|c|ccc|c}
		&  \textcolor{black}{Deterministic} & \textcolor{black}{Stochastic} & \textcolor{black}{Robust} & \textcolor{black}{Online} & \textcolor{black}{Total}\\ \hline
		\textcolor{black}{Level 1} & \textbf{57}  (56/1) & \textbf{72}   (48/24) & \textbf{ 5}  (5/0) & -- & \textbf{122}  (97/25)\\
                          Level 2  & \textbf{ 8}   (8/0) & \textbf{ 3}     (1/2) & \textbf{ 0}  (0/0) & -- & \textbf{ 11}    (9/2)\\
		\textcolor{black}{Level 3} & \textbf{89}  (83/6) & \textbf{107}  (55/52) & \textbf{14} (14/0) & -- & \textbf{198} (142/56)\\ \hline
		\textcolor{black}{Overall} & \textbf{147} (140/7) & \textbf{177} (101/76) & \textbf{19} (19/0) & -- & \textbf{319} (238/81)\\
	\end{tabular}
    \vspace{2mm}
	\caption{Absolute frequencies of different uncertainty modeling approaches overall and differentiated by level of integration. The numbers in parentheses correspond to the numbers of publications using/\emph{not} using optimization. Note that the sum of \rev{papers} in a \rev{row} may be larger than the overall number of publications \rev{in the column ``Total''} due to papers being counted twice (if two approaches are used). Similarly, the numbers for levels~1\rev{, level~2,} and~3 in a column may sum up to \rev{more} than the overall number in the bottom line \rev{due to papers being counted twice (if various levels are considered).}}
\label{tab:uncertainty-and-integration}
\end{table}

Concerning the frequencies of different uncertainty modeling approaches, it turns out that, for \rev{all integration levels and in particular for level~1 and level~3,} the vast majority of papers that consider uncertainty use stochastic approaches (\rev{177}), only a few use robust approaches (\rev{19}), while online \rev{optimization (competitive analysis)} approaches are not used at all. The absence of online \rev{optimization} approaches can potentially be explained by the fact that stochastic and robust modeling approaches stem from the field of OR/MS considered here, while online optimization has its origins in computer science~\cite{Bakker_2020}.

Another interesting observation is that almost all of the presented deterministic approaches are optimization models (\rev{140} of \rev{147}), while simulation approaches and other approaches almost always consider uncertainty in at least some of their input data. One reason for this could be that considering uncertain parameters is generally easier in simulation models than in optimization models.

While it is very common for simulation models to consider a large number of uncertain parameters simultaneously, a choice must often be made in optimization models to consider only a limited number of uncertain parameters. Therefore, we now analyze which parameters are most frequently considered as uncertain in optimization models.
Since the considered (uncertain) problem parameters naturally depend on the planned resources, however, we first analyze the resources that are planned within optimization models considering uncertainty. Here, we observe that most of the papers that consider uncertainty in optimization approaches \rev{look at} the OT~(\rev{77} papers), followed by physicians~(\rev{75}), beds~(\rev{60}), nurses~(\rev{40}), and ICU~(\rev{27}). Only few papers \rev{consider} the remaining resources, \rev{such as, e.g., medical equipment (7) or Diagnostics (6).} Interestingly, only very few of these papers consider only two resources -- the vast majority considers \rev{at least three resources. }
Turning to the analysis of which parameters are considered as uncertain, we observe that, when the OT is considered as a resource, the surgery duration is by far the most frequently used uncertain parameter in optimization models. Despite OT planning being a well-studied field of research, we identified only a limited number of papers that model factors other than the surgery duration to be uncertain. Among these uncertain parameters are arrivals, i.e., authors consider the number of arriving patients as uncertain~\citesearchresults{Astaraky_2015,Range_2019,Zhu_2022,Bansal_2021,Breuer_2020,Oostrum_2008,Zhang_2021,McRae_2020}. This may in some cases include the possibility of emergency arrivals as a separate source~\rev{\citesearchresults{Breuer_2020,Rachuba_2017,Zhu_2022,Jittamai_2011,Mazloumian_2022,Razmi_2015,Wan_2023,Wang_2023,Li_2023,Kamran_2019}}, which are otherwise only considered implicitly, e.g., as a proportion of the non-emergency arrivals or via OT time reservation~\citesearchresults{Molina-Pariente_2018}. Other uncertainty aspects such as no-shows~\citesearchresults{Jittamai_2011, Yan_2022}, patients reneging from waiting lists~\citesearchresults{Astaraky_2015} or cancellations of surgeries~\rev{\citesearchresults{Astaraky_2015,Augustin_2022,Kazemian_2017}}, are only considered a few times. When the OT is linked with up-/downstream resources, uncertainty in the length of stay is frequently considered. Unusual (i.e., not frequently modeled) uncertain parameters are the discharge rate of patients (from a hospital unit)~\citesearchresults{Zhu_2022}, the demand for beds (similar to uncertain arrivals)~\citesearchresults{Belien_2007,Kheiri_2021,Ma_2013}, or nurse or surgeon availability~\citesearchresults{Nasiri_2019,Breuer_2020,McRae_2020}. \rev{In particular, while many papers consider the integration of the OT with either medical staff or beds, only very few papers consider parameters other than durations to be uncertain -- even though unavailability of (medical) staff, for example, has already been identified as a highly relevant aspect within the (healthcare) personnel scheduling literature~\cite{Erhard+etal:phys-sched-survey,VandenBergh+etal:survey,Thielen_2018}.}
The paper by Hulshof et al.~\citesearchresults{Hulshof_2016} is the only one to consider the care pathway of patients to be uncertain. 
Lastly, it is interesting to note that papers using a robust approach to solve an OT-related problem unanimously focus on the surgery duration as the uncertain parameter~\citesearchresults{Bansal_2021,Breuer_2020,Neyshabouri_2017,Rachuba_2017,Rath_2017,Shehadeh_2021,Wang_2021,Keyvanshokooh_2022,Davarian_2022,Mazloumian_2022}. Very few of these papers consider other parameters to be uncertain, e.g., the need for surgery~\citesearchresults{Bansal_2021}, surgeon availability~\citesearchresults{Breuer_2020}, emergency arrivals~\citesearchresults{Breuer_2020,Rachuba_2017,Mazloumian_2022}, or length of stay in a downstream unit~\citesearchresults{Neyshabouri_2017,Shehadeh_2021}. 
Within the papers in which the OT is \emph{not} considered as a resource, we found that the arrival rate of patients (or the number of patients, including one paper modeling no-shows)~\citesearchresults{Augusto_2009,He_2019,Izady_2021,Kortbeek_2017,Leeftink_2019,Agrawal_2023,Gong_2022,Chan_2022,Kobara_2022,Marchesi_2020,Schäfer_2019,Schäfer_2023,Rahiminia_2023,Andersen_2019,Yan_2022,Allihaibi_2021,Daldoul_2022,Zhou_2022_a}, treatment \rev{or consultation times}~\citesearchresults{He_2019,Gong_2022,Haghi_2022,Kobara_2022,Dehghanimohammadabadi_2023,Gul_2023,Rahiminia_2023,Allihaibi_2021,Daldoul_2022,Demir_2021,Xiao_2017}, and bed demand~\citesearchresults{Ordu_2021,Schoenfelder_2020} are commonly-used uncertain parameters. Again, in almost all cases, the uncertain parameters are patient- or demand-related, while one of these publications investigates pharmacies inside hospitals and considers the delivery of medicines as uncertain~\citesearchresults{Augusto_2009}.

\section{Practical implementation and data}\label{sec:data-and-implementation}
We also scanned our search results for information on practical implementation and the use of real data. Here, we observe two principal ways in which methods or results are used in practice. The first way is that the corresponding paper makes a suggestion for a one-time change in practice that is subsequently implemented at one or several partner hospitals. For example, Toronto’s Mount Sinai Hospital eliminated its Thoracic Surgery service during budget negotiations based on an advice from Blake and Carter~\citesearchresults{Blake_2002}, and recommendations made by Kortbeek at al.~\citesearchresults{Kortbeek_2017} based on their research on the trade-offs between appointment scheduling constraints and access times were implemented in the Academic Medical Center in Amsterdam.
The second way in which methods and results are used in practice is that the corresponding paper present a decision support tool that is then used on a regular basis to solve recurring integrated planning problems in hospitals. For example, the integrated surgery scheduling approach developed by Ozen et al.~\citesearchresults{Ozen_2016} was implemented as a web-based application and integrated into the existing surgical planning systems at Mayo Clinic.

Overall, only \rev{31} of the \rev{319} relevant papers that resulted from our search report an actual application of their work in practice in one of the above-mentioned ways. In contrast, \rev{207} papers present a case study without mentioning a practical application, and \rev{80}~papers have a primarily methodological focus, i.e., they describe a new model or solution approach that is not directly connected to a case study or a practical application. Among the papers that do mention a practical application of their results, the earliest one was published already in 2002 on strategic resource allocation in acute care hospitals~\citesearchresults{Blake_2002}.

\begin{figure}
	\centering
	\includegraphics[width=0.99\linewidth]{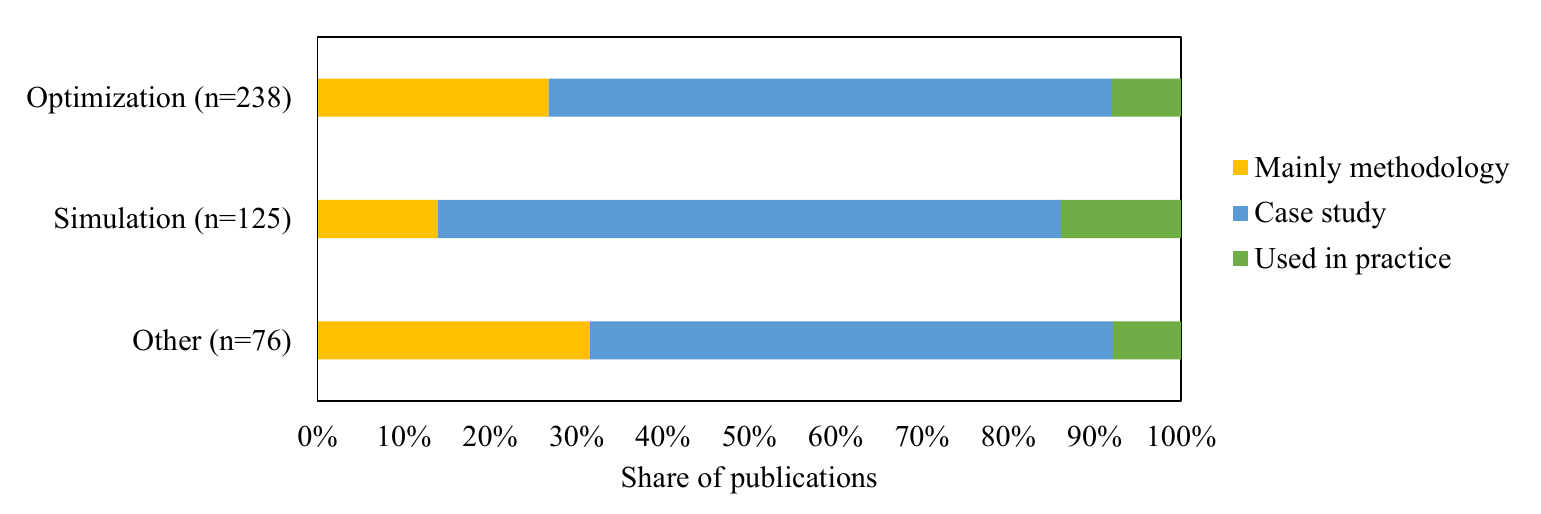}
	\caption{Degree of implementation across methodological focus areas.}
	\label{fig:methods-and-implementation}
\end{figure}

Analyzing the degree of practical application depending on the utilized modeling and solution methods as shown in Figure~\ref{fig:methods-and-implementation}, we observe that the highest share of publications whose results are used in practice is found among those using simulation methods, while the share of papers that are mainly methodologically focused is by far the lowest among papers using simulation. Among both the papers using optimization methods and those using other methods, the share used in practice is much lower, while papers with a methodological focus have a much higher relative frequency. Throughout all three categories of methods, however, the vast majority of papers present case studies \rev{without reporting subsequent} practical applications of the developed methods and results. \rev{This is in line with the findings of previous literature reviews that -- while not focusing specifically on integrated planning -- also found little evidence for successful implementation of research output~\cite{Brailsford2009:simulation-survey,Cardoen:operating-room-survey, Erhard+etal:phys-sched-survey}.}

Concerning the use of real-world data, it is not surprising that \rev{30} of the \rev{31}~publications reporting practical implementations use real data, while the remaining one at least uses realistic data, i.e., artificial data that has been validated as realistic through discussions with practitioners and / or literature research. Given the low overall number of \rev{31}~papers in which the authors report practical applications of their work, it is more surprising that the vast majority of the papers in our search results (\rev{221} of \rev{319}) still report on the use of real data. Together with the above observation that the number of case studies vastly exceeds the number of papers that have lead to practical applications, this suggests that obtaining real data on integrated planning problems in hospitals to use in a case study is significantly easier than actually bringing the results of a research project from this area into practice. Moreover, even the \rev{31}~papers that report practical applications of their methods or results mostly provide only brief descriptions about practical impact and / or implementation in the publications themselves in one or two paragraphs at the end of the paper. This \rev{indicates} that scientific journals and conferences that publish work on integrated planning in hospitals do not put much emphasis on \rev{generating impact using} the obtained results so far.

While cross comparisons of the obtained degree of practical application and the level of integration or the considered primary resources do not yield any significant insights, another noteworthy observation is that \rev{almost} all of the publications whose results on integrated planning are used in practice report about practical applications in Europe or the Americas, while \rev{almost} no transfers into practice are reported in the rest of the world. 

\rev{\section{State-of-the-art}\label{sec:state-of-the-art}}
\rev{Due to the variety of resource combinations and methods that are used in the relevant papers found in our search, an in-depth discussion of all search results has limited use. Instead, this section briefly discusses and compares 59~recent publications (published since 2021) that are classified as completely integrated planning (level~3) according to our taxonomy to identify the most recent research trends. The discussion is organised by the primary resource and then further subdivided according to hierarchical planning levels.}

\rev{\subsection{Operating theater (OT)}\label{subsec:SOTA_OT}}
\rev{\paragraph{Strategic:} The only paper that uses a System Dynamics model to support strategic decision making in OTs including physician capacity is~\citesearchresults{Faeghi_2021}. The authors investigate the long-term responsiveness of operating theaters and consider surgical departments and physicians as producing units. Operating theater capacity and recovery beds are considered in~\citesearchresults{Bavafa_2022} and~\citesearchresults{Kheiri_2021}. In both cases, a stochastic optimization model is developed to solve the underlying problem. While ~\citesearchresults{Bavafa_2022} focuses on how effective coordination of case-mix decisions and discharge strategies can impact the utilization of operating theaters and recovery beds, \citesearchresults{Kheiri_2021} aims to avoid cancellations in surgery schedules.}

\medskip

\rev{Besides these purely strategic approches, there are also approaches that combine strategic decision making with tactical and/or operational decisions. 
A combination of strategic and tactical levels supporting master surgery scheduling is used in~\citesearchresults{Kianfar_2023}, where two types of full-day and half-day blocks are constrained by available beds on inpatient wards. Strategic, tactical, and operational decisions when simultaneously planning operating theaters and surgeons are considered via three optimization models using lexicographic objectives in~\citesearchresults{Akbarzadeh_2020}. Since a stochastic approach is considered intractable, a proactive-reactive hierarchical approach is used with an iterative combination of simulation and optimization.}

\newpage

\rev{\paragraph{Tactical:}}
\rev{On the tactical level, master surgery schedules~(MSS) (e.g., \citesearchresults{Heider_2022_a,Deklerck_2021,Makboul_2022_a}) or similar forms of regularly usable blueprints~\citesearchresults{Rachuba_2022} for capacity allocation are combined with allocation of beds (for inpatient wards or ICU) or medical staff. Focusing initially on purely tactical approaches (i.e., not links to strategic or operational planning), all publications combine OT and bed capacity, which can be for inpatient wards \citesearchresults{Zhu_2022,Carter_2023,Bovim_2022}, ICU \citesearchresults{Makboul_2022_a,Rachuba_2022}, PACU \citesearchresults{Makboul_2022_a,Zhu_2022}, or specified as a general \emph{bed} or \emph{downstream resource} \citesearchresults{Zhou_2022,Lazebnik_2023,Heider_2022_a}. Nurses are only explicitly considered in~\citesearchresults{Lazebnik_2023,Heider_2022_a}, while other papers state that beds are meant as staffed beds~\citesearchresults{Bovim_2022}. Physicians are, in turn, explicitly mentioned in \citesearchresults{Deklerck_2021,Zhou_2022,Bovim_2022,Carter_2023,Arab_Momeni_2022,Lazebnik_2023,Makboul_2022_a,Heider_2022_a}, while~\citesearchresults{Rachuba_2022,Zhu_2022} do not consider them. It is interesting to note that~\citesearchresults{Bovim_2022} is the only paper that links the OT to outpatient clinic appointments. In a combination of tactical and operational planning, \citesearchresults{Akbarzadeh_2023} combines OT and physicians, while~\citesearchresults{Moustafa_2022,Heider_2022_a} include phycisians and also different bed types (e.g., inpatient ward, ICU).}

\medskip

\rev{The only approach spanning strategic, tactical, and operational decision making is~\citesearchresults{Akbarzadeh_2023}, in which an integrated optimization model is used for strategic and tactical planning, which is also connected to separate models for operational offline and operational online planning. 
Strategic and tactical OT planning are considered in~\citesearchresults{Kianfar_2023}. Operational consequences of tactical planning are considered as a joint MSS and surgical case scheduling approach in~\citesearchresults{Moustafa_2022}, while \citesearchresults{Heider_2022_a,Rachuba_2022,Deklerck_2021} use discrete event simulation models to exemplify the use of tactical blueprints over time. In general, many studies formulate deterministic optimization models \citesearchresults{Deklerck_2021,Rachuba_2022,Bovim_2022,Kianfar_2023,Carter_2023,Akbarzadeh_2023,Moustafa_2022}, whereas only~\citesearchresults{Lazebnik_2023,Heider_2022_a} use stochastic optimization and~\citesearchresults{Arab_Momeni_2022,Makboul_2022_a} use a robust optimization approach. A unique combination of methods is used in~\citesearchresults{Lazebnik_2023}, where agent-based simulation is combined with deep reinforcement learning to maximize treatment success. Hence, this is also the only study that does not use optimization techniques to determine an MSS, a tactical blueprint, or a schedule.}

\rev{\paragraph{Operational (offline \& online):}}
\rev{With a focus on the OT, operational offline decisions are usually concerned with selecting patients from a waiting list and assigning them to operating rooms and days of the week. These decisions also include the sequencing of patients on a given day and/or in a given operating room~\cite{Hulshof:taxonomy}.}

\rev{We initially focus on purely operational decisions that are not linked to other hierarchical levels. The majority of theses papers links OT scheduling decisions with physicians \citesearchresults{Naderi_2021,Wang_2021,Keyvanshokooh_2022,Wan_2023,Augustin_2022,Azar_2022,Gaon_2023}. Other studies link the OT to PACU beds~\citesearchresults{Lu_2021}, ICU beds~\citesearchresults{Heider_2022}, or ambulance dispatching during mass casualty incidents~\citesearchresults{Zhu_2023}. While~\citesearchresults{Heider_2022} aims to minimize peak utilizations in the ICU using a quota system for the number of elective surgeries on a day, \citesearchresults{Zhu_2023} aims to maximize the number of patients that can undergo surgery in case of a large-scale emergency, where the decisions concern selecting patients for surgery and sequencing of surgeries. A more general perspective is found in~\citesearchresults{Lu_2021}, where \emph{resources for surgery scheuduling} are considered. Stochastic \citesearchresults{Lu_2021,Spratt_2021_b,Keyvanshokooh_2022,Wan_2023,Augustin_2022,Azar_2022,Schoenfelder_2021} as well as robust \citesearchresults{Wang_2021,Keyvanshokooh_2022,Davarian_2022} approaches are frequently used on this decision level. Uncertainty is considered for activity durations and demand.}

\medskip

\rev{Selecting patients and sequencing them, while explicitly considering physicians, is handeled in two separate optimizations models in~\citesearchresults{Akbarzadeh_2023}; one for operational offline and one for operational online planning. The overall approach connects these two models to a third model that supports strategic and tactical decision making. Further tactical and operational offline connections are made in~\citesearchresults{Moustafa_2022,Heider_2022_a}. In~\citesearchresults{Moustafa_2022}, surgery scheduling is an optimization-based extension to the MSS, whereas~\citesearchresults{Heider_2022_a} employs a simulation model to evaluate the operational impact of the proposed MSS. Both approaches include decisions on physicians and inpatient beds.}

\medskip

\rev{Focusing on surgery scheduling after a disaster, \citesearchresults{Li_2023} is the only paper in this subsection's selection that considers an \emph{operational online} problem. The optimization model assigns elective and emergency surgeries to operating rooms, sequences them, and assigns patients to beds for anesthesia and recovery with the aim of simultaneously minimizing the expected recovery completion time for all patients
(makespan), the expected average makespan while prioritizing emergency / urgent patients, as well as the
expected total idle time of operating rooms.}

\rev{\subsection{Physicians}\label{subsec:SOTA_physicians}}
\rev{\paragraph{Strategic:} Purely strategic problems are tackled in \citesearchresults{Faeghi_2021,Ordu_2021,Rise_2021}. The System Dynamics approach in \citesearchresults{Faeghi_2021} allows to investigate long-term effects of the dynamics with respect to surgery planning in an OT system while also considering surgeons. The focus of \citesearchresults{Ordu_2021,Rise_2021}, in turn, is on an entire hospital. Combining forecasting, simulation, and optimization, \citesearchresults{Ordu_2021} aims to improve bed and staff allocation. A combination of qualitative research and stochastic optimization  with multiple objectives is used in \citesearchresults{Rise_2021} to decide on the number of resources needed for each department in a hospital.}

\rev{A strategic-tactical approach is presented in \citesearchresults{Kianfar_2023}, where OT decisions are combined with the number and allocation of surgeons within a multiobjective MILP. Spanning further across hierarchical levels, \citesearchresults{Dosi_2021,Akbarzadeh_2023} consider strategic, tactical, and operational problems. Optimization models for different decision levels are connected in~\citesearchresults{Akbarzadeh_2023} to link OT and physician planning, while~\citesearchresults{Dosi_2021} uses a combination of design thinking and discrete event simulation to support nurse and physician planning in an ED.}

\rev{\paragraph{Tactical:} On the tactical level, physician planning is combined mainly with OT \citesearchresults{Deklerck_2021,Bovim_2022,Arab_Momeni_2022,Lazebnik_2023,Makboul_2022_a} or nurse planning \citesearchresults{Deklerck_2021,Kuo_2023,Atalan_2022,Battu_2023,Lazebnik_2023,Corsini_2022}, and~\citesearchresults{Deklerck_2021,Lazebnik_2023} consider all three of them together. Another interesting focus is that of~\citesearchresults{Kuo_2023}, where a bed and medical staff allocation problem is modeled using a multiobjective simulation-based optimization model. Similarly, yet without beds as a resource, \citesearchresults{Atalan_2022,Battu_2023} aim to improve patient flows in EDs. In~\citesearchresults{Atalan_2022}, different configurations of medical staff are evaluated, while~\citesearchresults{Battu_2023} also includes medical imaging devices and paramedical staff. Lastly, with a focus on chemotherapy, \citesearchresults{Corsini_2022} consider medical staff (nurses and physicians) and in addition pharmacy technicians.}

\rev{Tactical-operational decisions focus, for example, on improving ED performance using a simulation-based optimization approach~\citesearchresults{Daldoul_2022}. Another approach extends an existing MSS with the aim to minimize maximimum workload for medical staff and evaluates the obtained MSS with a simulation model~\citesearchresults{Heider_2022_a}. }

\rev{\paragraph{Operational (offline \& online):} On the purely operational offline level, OT-related physician planning problems are considered in \citesearchresults{Naderi_2021,Wang_2021,Keyvanshokooh_2022,Wan_2023,Augustin_2022,Azar_2022,Gaon_2023}. Of those, only~\citesearchresults{Gaon_2023} considers nurses, physicians, and OT together, which is done via multi-agent optimization models that aim to maximize the overall utility of all agents. On the other hand, operational models that do not include the OT focus on chemotherapy and its dedicated medical equipment~\citesearchresults{Bouras_2021,Haghi_2022}, ED planning \citesearchresults{Taleb_2023,Allihaibi_2021,Queiroz_2023}, or laser treatment in an ophtalmology department~\citesearchresults{Kanoun_2023}. The distribution of patients, doctors, and medical equipment in a health network of several hospitals is considered in~\citesearchresults{Gutierrez_2021} with the objective of meeting daily demand while minimizing access times and patient transfers. On this level, deterministic \citesearchresults{Bouras_2021,Naderi_2021,Wang_2021,Gutierrez_2021,Kanoun_2023,Queiroz_2023} as well as stochastic \citesearchresults{Spratt_2021_b,Keyvanshokooh_2022,Haghi_2022,Wan_2023,Augustin_2022,Allihaibi_2021,Azar_2022} and robust~\citesearchresults{Wang_2021,Keyvanshokooh_2022} optimization approaches are used. 
Simulation techniques are less frequently applied \citesearchresults{Taleb_2023,Allihaibi_2021,Azar_2022,Schoenfelder_2021}.}

\rev{On the combined operational \emph{offline} and \emph{online} level, two publications focus only on this planning level~\citesearchresults{Spratt_2021_b,Schoenfelder_2021}. An MSS with a rolling time horizon that maximizes the scheduled patient priorities is used in \citesearchresults{Spratt_2021_b}, whereas a simulation study is used to evaluate management strategies for the OT in~\citesearchresults{Schoenfelder_2021}. In both cases, surgeon assignment is part of the planning decisions.}

\rev{Operational \emph{online} physician planning in the context of operating theaters is considered in~\citesearchresults{Schoenfelder_2021,Akbarzadeh_2023,Spratt_2021_b}. While the focus of~\citesearchresults{Schoenfelder_2021} and~\citesearchresults{Akbarzadeh_2023} is on evaluating planning and scheduling policies in an OT, \citesearchresults{Spratt_2021_b} presents a rolling horizon approach over a number of weeks that links strategic, tactical, and operational decision making. In~\citesearchresults{Yan_2022}, patient preferences for physicians and appointment times are considered.}

\rev{\subsection{Beds}\label{subsec:SOTA_beds}}
\rev{\paragraph{Strategic:} On this level, bed planning is usually concerned with dimensioning and determining the number of required beds. In~\citesearchresults{Izady_2021} and \citesearchresults{Kobara_2022}, dedicated hospital wards are dimensioned. While~\citesearchresults{Kobara_2022} balances bed capacity in the ICU and a step-down unit, \citesearchresults{Izady_2021} combines beds of different medical departments into a clustered overflow configuration and also considers staff availability (e.g., via nurse-to-patient-ratios). Forecasting and managing patient demand for an entire hospital are considered in~\citesearchresults{Ordu_2021,Rise_2021}. In~\citesearchresults{Ordu_2021}, a hybrid simulation-based framework is used to predict and optimize the allocation of beds and staff. For similar decisions, \citesearchresults{Rise_2021} use a combined qualitative and quantitative approach. In two further papers~\citesearchresults{Bavafa_2022,Kianfar_2023}, beds are considered as downstream resources of the OT, with a focus on case-mix decisions and different block types in an MSS, respectively.}

\rev{\paragraph{Tactical:} On the tactical level, beds are frequently considered as a downstream resource in combination with OT planning (e.g., in \citesearchresults{Bovim_2022,Kianfar_2023,Carter_2023,Arab_Momeni_2022}). Two further possible links include the ED~\citesearchresults{Kuo_2023,Daldoul_2022} and a broader view on resource planning in general~\citesearchresults{Zhou_2022,Lazebnik_2023}. Of the OT papers, \citesearchresults{Carter_2023} focuses on leveling the utilization of inpatient wards, while~\citesearchresults{Kianfar_2023,Bovim_2022,Arab_Momeni_2022} also include decisions on medical staff. For ED planning, \citesearchresults{Kuo_2023} and~\citesearchresults{Daldoul_2022} include medical staff and beds in a simulation-optimization approach and an MIP with subsequent DES evaluation, respectively. An agent-based approach to allocate hospital staff and other resources (OTs, medication / drugs, and diagnosis machines) is proposed in~\citesearchresults{Lazebnik_2023}, while~\citesearchresults{Zhou_2022} uses two-stage stochastic optimization for multi-resource planning with multi-type patient planning.}

\rev{\paragraph{Operational (offline \& online):} On the operational offline level, bed planning is linked to ED~\citesearchresults{Boyle_2022,Allihaibi_2021,Daldoul_2022}, OT and ICU~\citesearchresults{Heider_2022,Wan_2023}, ICU and regular wards~\citesearchresults{Gokalp_2023}, and infusion appointments~\citesearchresults{Issabakhsh_2021}.  A generic DES model to assess changes in the ED is presented in~\citesearchresults{Boyle_2022}, and~\citesearchresults{Allihaibi_2021} develops a simulation-optimization approach to design an emergency care patient pathway. In turn, \citesearchresults{Daldoul_2022} decides on the setup for the entire ED, including the numbers of physicians, nurses, and beds, while~\citesearchresults{Heider_2022} uses a quota system for OT scheduling to minimize peaks in ICU bed occupancy. Similarly, \citesearchresults{Wan_2023} combines OT scheduling with decisions on the number of beds in the ICU, and also includes surgeons' preferences for working days.}

\rev{On the operational online level, \citesearchresults{Lee_2021} coordinates inpatient admissions from the ED based on a queueing model, while~\citesearchresults{Li_2023} supports post-disaster scheduling of as part of the peri-operative process for elective and emergency patients.}

\rev{\subsection{Nurses}\label{subsec:SOTA_nurses} In general, only few publications consider nurses individually \citesearchresults{Chan_2022,Gul_2023,Demir_2021} and not in direct connection to physicians as in \citesearchresults{Bouras_2021,Dosi_2021,Ordu_2021,Rise_2021,Haghi_2022,Taleb_2023,Kuo_2023,Allihaibi_2021,Atalan_2022,Battu_2023,Daldoul_2022,Lazebnik_2023,Schoenfelder_2021,Corsini_2022,Heider_2022_a}.}
\rev{\paragraph{Strategic:} On the strategic level, \citesearchresults{Dosi_2021} plans nurses and physicians using a DES model and design thinking approaches for an ED. Nurses as part of a whole-hospital planning concept based on a forecasting-simulation-optimization approach are the focus of~\citesearchresults{Ordu_2021}. Another overall perspective on the hospital is considered in~\citesearchresults{Rise_2021}, but here, the authors study how changes in one department impact another one using a simulation-based multiobjective optimization approach. Imaging services such as CT or MRT are explicitly included in~\citesearchresults{Dosi_2021} and~\citesearchresults{Rise_2021}.}

\rev{\paragraph{Tactical:} All eight papers using tactical approaches with a focus on nurses also consider physicians. Most of them are applied in an ED context, focusing on overcrowding and patient flows~\citesearchresults{Dosi_2021}, allocation of medical resources (staff and also diagnostic) using DES models~\citesearchresults{Kuo_2023,Atalan_2022}, or combined simulation-optimization approaches~\citesearchresults{Battu_2023,Daldoul_2022}. A hospital-wide approach in~\citesearchresults{Lazebnik_2023} uses an agent-based simulation approach and reinforcement learning. An MSS extension to level ICU beds is presented in~\citesearchresults{Heider_2022_a}, and~\citesearchresults{Corsini_2022} address the design of an outpatient chemotherapy department using simulation. Not only nurses and physicians, but also support staff (e.g., laboratory or pharmacy technicians) are considered in~\citesearchresults{Atalan_2022,Battu_2023,Corsini_2022}. Only~\citesearchresults{Dosi_2021} considers medical imaging as part of this planning level.} 

\rev{\paragraph{Operational (offline \& online):} Some of the previously mentioned approaches also consider operational offline decisions~\citesearchresults{Dosi_2021,Daldoul_2022,Heider_2022_a}. Of the remaining ones, only~\citesearchresults{Gul_2023,Demir_2021} focus solely on nurses without including physicians, and~\citesearchresults{Bouras_2021,Haghi_2022} take medical equipment for chemotherapy into consideration, while the paper~\citesearchresults{Schoenfelder_2021} is the only one to link OT planning to nurses and physicians. In addition to physicians and nurses, \citesearchresults{Taleb_2023,Allihaibi_2021} focus on EDs and include receptionists and triage staff, respectively.} 

\rev{On the operational online level, \citesearchresults{Chan_2022} re-assigns two types of nurses to areas in an ED at the beginning of a shift, and, among other resources,~\citesearchresults{Schoenfelder_2021} also considers nurses.}

\rev{\subsection{Appointments/Admissions}\label{subsec:SOTA_appointments}}
\rev{\paragraph{Strategic:} An appointment system for surgery planning that considers downstream resources and also integrates tactical decisions is designed in~\citesearchresults{Kianfar_2023}. The paper~\citesearchresults{Rahiminia_2023} has a unique focus and simulatneously considers congestion management and medical waste reduction, which is modeled using a Markovian queueing-based inventory system.}

\rev{\paragraph{Tactical:} On the tactical level, \citesearchresults{Bovim_2022,Kianfar_2023} simultaneously consider patient appointments, OT capacity, medical staff, and beds. The paper~\citesearchresults{Zhou_2022} has a similar focus, but uses a generic approach for any type of resource instead of explicitly mentioned resources. As another recent contribution, \citesearchresults{Bentayeb_2023} integrates patient appointments with schedules of technicians at a radiology center.}

\rev{\paragraph{Operational (offline \& online):} On the operational offline level, \citesearchresults{Bouras_2021,Haghi_2022,Gul_2023,Demir_2021} link appointment planning decisions in chemotherapy planning to medical equipment and also to physicians and nurses. In addition, \citesearchresults{Haghi_2022,Gul_2023} consider stochastic treatment or infusion times, \citesearchresults{Demir_2021} includes stochasticity in durations for premedication and infusion, while~\citesearchresults{Bouras_2021} uses a deterministic planning approach. Combinations of physicians and appointments are considered for surgical care~\citesearchresults{Keyvanshokooh_2022}, for dynamic scheduling in EDs~\citesearchresults{Queiroz_2023}, and for an infusion center using a robust optimization approach~\citesearchresults{Issabakhsh_2021}.}

\rev{The only operational online paper~\citesearchresults{Yan_2022} schedules appointments for patients in an outpatient clinic with different physicians and considers probabilities that patients accept an offer.}

\rev{\subsection{ED}\label{subsec:SOTA_ED}}
\rev{\paragraph{Strategic:} In~\citesearchresults{Dosi_2021,Rise_2021}, ED planning is combined with decisions on medical staff and medical imaging devices (CT and MRT) are also included. In a more holistic approach, \citesearchresults{Rise_2021} further considers inpatient or ICU beds, treatment rooms, the OT, and additional staff categories. While~\citesearchresults{Dosi_2021} uses a DES model, \citesearchresults{Rise_2021} employs a simulation-optimization approach.}

\rev{\paragraph{Tactical:} The above-mentioned paper~\citesearchresults{Dosi_2021} considers strategic, tactical, and operational offline decisions simultaneously. A discrete event simulation is presented in~\citesearchresults{Atalan_2022,Battu_2023} that consider nurses, physicians, and support staff. In~\citesearchresults{Battu_2023}, a medical MRI as well as ultrasound and X-ray devices are additionally considered. Lastely, in~\citesearchresults{Kuo_2023,Daldoul_2022}, simulation-based optimization approaches are used to allocate staff and beds.}
\rev{\paragraph{Operational (offline\&online):} ED processes and bed capacity within a DES model that allows to analyze various setups are considered in~\citesearchresults{Boyle_2022}. In contrast, \citesearchresults{Allihaibi_2021,Daldoul_2022} use simulation-optimization approaches to minimize patient waiting times and the overall length of stay. A unique study is~\citesearchresults{Taleb_2023}, where the performance of an ED is evaluated using a combination of DES and Data Envelopment Analysis. The only paper in this selection not considering staff is~\citesearchresults{Boyle_2022}, while~\citesearchresults{Dosi_2021,Taleb_2023} do not consider allocation of bed capacity. No operational \emph{online} papers with the ED as a leading resource remain after filtering for this section.}

\section{Integration with other parts of a healthcare system}\label{sec:hospital+outside}
Since hospitals usually have dependencies with many other care providers and healthcare services, this section provides an outlook on existing planning approaches that link a hospital to other hospitals and other parts of a healthcare system. 

From a patient's point of view, hospitals are one part of their overall care pathway. For example, emergency patients might have been taken to the emergency department of a hospital by an ambulance. Elective patients have been diagnosed and potentially treated previously by general practitioners or specialists and have then been transferred to a hospital for further treatment. During their hospital stay, they might need medication or blood bags that must be ordered and delivered, or they might need to be transferred from one hospital to another. Afterwards, patients may receive follow-up care at home or are transferred to a rehabilitation facility. 

\smallskip

In the following, we distinguish three kinds of dependencies of hospitals with other entities in a healthcare system based on the position of these entities on a patient's care pathway:
\begin{enumerate}
	\item \emph{Pre-hospital dependencies:} Dependencies with care providers or healthcare services that are positioned before a hospital stay on a patient's care pathway,
	\item \emph{During-hospital dependencies:} Dependencies with other hospitals, care providers, or healthcare services (e.g., blood banks) that are relevant during a patient's hospital stay,
	\item \emph{Post-hospital dependencies:} Dependencies with care providers or healthcare services that are positioned after a hospital on a patient's care pathway.
\end{enumerate}

\begin{figure}[!ht]
	\centering
	\includegraphics[width=\textwidth]{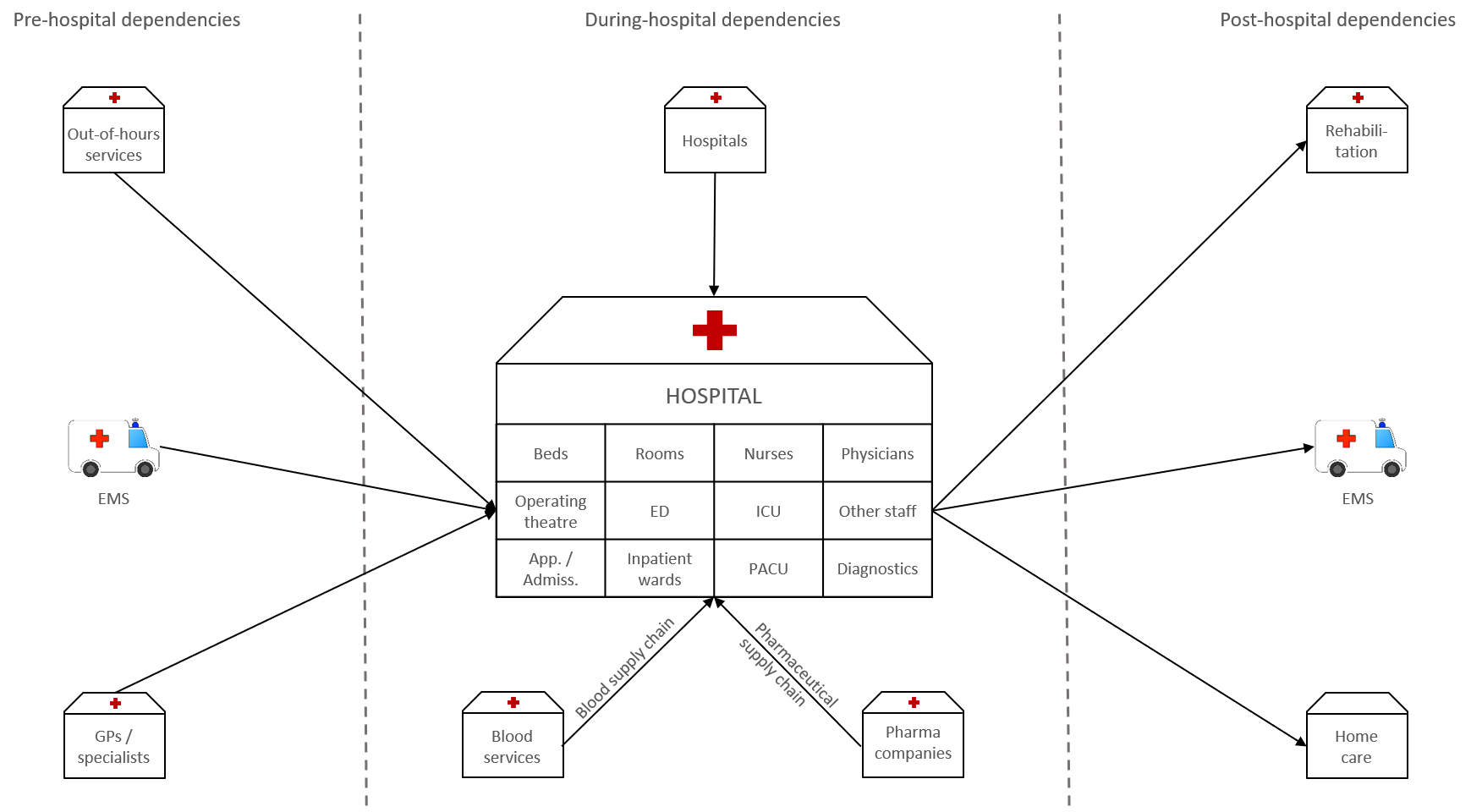}\\[1em]
	\caption{Overview of dependencies of a hospital within a healthcare system.}
	\label{fig:System}
\end{figure}

An overview of dependencies of a hospital within a healthcare system that distinguishes between pre-, during-, and post-hospital dependencies is shown in Figure~\ref{fig:System}.
For many of these dependencies, a simultaneous consideration and an integrated planning of the involved resources can be beneficial from both a patient and a system perspective and can even improve individual objectives for the involved care providers. In the following, we look more closely at three examples \rev{(one for each case) of interactions of hospitals with other parts of a healthcare system} that have already been addressed in the OR/MS literature. \rev{We provide a general overview of each exemplary interaction by providing pointers to the existing literature, and we identify possible directions for future research.}

\subsection{Pre-hospital dependencies: Ambulance diversion and offload delay}
While hospitals naturally interact at least indirectly with most entities that are usually positioned before a hospital stay on a patient's care pathway (e.g., general practitioners), one of the most-studied and most direct interdependencies is between emergency departments (EDs) of hospitals and emergency medical services~(EMS). When an ED is crowded, ambulances might need to be diverted to other hospitals, which is referred to as \emph{ambulance diversion}. Alternatively, they might wait in front of the hospital until patients can be admitted to the ED, which leads to a so-called \emph{(ambulance) offload delay}. As defined in Li et al.'s literature review on offload delay~\cite{Li_2019}, ``ambulance offload delay (AOD) occurs when care of incoming ambulance patients cannot be transferred immediately from paramedics to staff in a hospital emergency department.''
Optimal control policies for ambulance diversion as a countermeasure to avoid offload delay are, for example, proposed by Ramirez-Nafarrate et al.~\cite{Ramirez_2014}. Besides others, Allon et al.~\cite{Allon_2013} study the impact of hospital size and occupancy on ambulance diversion in the US, and the effects of ambulance diversion are reviewed by Pham et al.~\cite{Pham_2006}.

So far, the OR/MS literature mainly addresses the ambulance diversion and offload delay problems from only one side, either focusing on the EMS or the ED, even though an exchange of information between EMS and ED together with a system-wide perspective is crucial in order to enable integrated planning~\cite{Reuter-Oppermann_2020}.

\subsection{During-hospital dependencies: Inter-hospital collaboration}
Many different kinds of dependencies of hospitals with other kinds of care providers and healthcare services are relevant during a patient's hospital stay and are therefore investigated in the OR/MS literature. Important examples include interactions with other entities in specific parts of a hospital's supply chain such as the blood supply chain~\cite{Belien_2012,Osorio_2015,Piraban_2019} or the pharmaceutical supply chain~\cite{Nguyen_2022}. 

Another important aspect that we focus on here and that relates directly to hospital resources is inter-hospital collaboration in the form of resource sharing. This means that different hospitals collaborate by sharing expensive resources such as imaging devices in order to gain efficiency. One main advantage of this type of inter-hospital collaboration is that ``hospitals can avoid the purchase of expensive medical resources and patients can be treated in a timely manner in any available hospital, which will improve their quality of care.''~\cite{Chen_2017}. Ideally, this leads to an integrated planning of the shared resources. Since resources such as imaging devices are usually not portable, one of the most crucial aspects of this is to plan how patients should be referred between the collaborating hospitals when the shared resources are required for their treatment. This question has, for example, been studied in~\cite{Chen_2013,Chen_2016,Chen_2017}. Chen and Juan~\cite{Chen_2013} consider the problem of daily patient referrals for CT scans between three hospitals, while Chen et al.~\cite{Chen_2016} and Chen and Lin~\cite{Chen_2017} investigate referring patients between two or multiple cooperating hospitals, respectively, that share imaging services. 

A related kind of hospital collaboration that also leads to patient referrals is utilization leveling among multiple hospitals with the goal of reducing disparities between the involved hospitals utilization rates. For example, Li et al.~\cite{Li_2018} study when patients should be referred from a high-utilization hospital to a low-utilization hospital and Li et al.~\cite{Li_2020} extend this investigation to a network of one high-utilization hospital and three low-utilization hospitals. Nezamoddini and Khasawneh~\cite{Nezamoddini_2016} also integrate capacity allocation decisions by determining the optimal resource levels of EDs in several hospitals while considering patient transfers between them.

Overall, while the question of patient referrals between resource-sharing hospitals or hospitals with different utilization rates has already been investigated as shown above, \rev{we did not find} much existing work that considers other aspects of integration that could potentially be relevant in settings in which equipment such as imaging devices is shared between hospitals.

\subsection{Post-hospital dependencies: Bed blocking}
Hospitals also interact in various ways with care providers that treat patients after their hospital stay. The perhaps most-studied aspect of these interactions is bed blocking in hospitals, which occurs when patients in a hospital are ready to be discharged but have to remain in the hospital until a bed in a follow-up care facility (e.g., in rehabilitation center or nursing home) becomes available~\cite{Rubin_1975,Zychlinski_2020}. Bed blocking can not only be harmful for patients due to the delay in advancing to the next step of their care pathway, but is also often costly since a hospital bed is more expensive to operate than, e.g., a \rev{bed in a geriatric institution}~\cite{Zychlinski_2020}. The problem of bed blocking has been studied intensively in the literature (see, e.g.,~\cite{El_1998,Mur_2011,Chemweno_2016,Rashwan_2015,Dong_2022}) and it has been recognized that better integration and cooperation between hospitals and follow-up care facilities is necessary in order to prevent it. For instance, Mur-Veeman and Govers~\cite{Mur_2011} state in their work on buffer management in Dutch hospitals to solve bed blocking that ``although stakeholders recognize that cooperation is imperative, they often fail to take the actions necessary to realize cooperation.'' Motivated by the improved integration that is needed in order to solve the bed blocking problem, Chemweno et al.~\cite{Chemweno_2016} model the complete care pathway for stroke patients to analyze the effects of different intervention strategies aimed at minimizing patient waiting time delays for available bed resources. Using simulation, they show that maximizing the bed resource utility leads to a decrease in patient waiting times. Rashwan et al.~\cite{Rashwan_2015} use system dynamics for obtaining a holistic and strategic national level capacity-planning model to address the problem of acute bed blocking in the Irish healthcare system, while Wood and Murch~\cite{Wood_2020} address the general problem of blocking after any service along the patient pathway using a continuous-time Markov chain.

In summary, bed blocking has already received considerable attention in the literature. In addition, the necessity for integrated models that address the problem by looking at all involved entities along the relevant care pathways has already been recognized. Nevertheless, such integrated models still seem to be scarce.

\newpage

\rev{\section{Research gaps, limitations, and conclusion}\label{sec:conclusion}}

\rev{This section identifies overarching trends and open research areas, discusses limitations of this review, and concludes our analyses with several take-home messages.}

\medskip

\subsection{Research gaps and open areas}
Despite the large number of publications that meet our inclusion criteria, the focus in terms of hospital areas and resources is still rather narrow. It becomes clear that there is a shortage of non-OT-focused publications, which might be due to the \emph{popularity} (or importance) of the OT itself or a lack of effort to explore other areas. A stronger focus on staff could be particularly promising here since staff shortages in hospitals are already visible and are expected to intensify in the future as population ageing increases demand for care relative to the size of the healthcare workforce. While staff \rev{are} already the most considered supplementary resource overall in integrated planning problems, we therefore expect also the center of attention to shift from the still mostly OT-focused planning observed so far to workforce-focused planning as a driver of future research. Moreover, supply services such as sterilization or pharmacy have not yet been included in integrated planning approaches. We therefore believe that increasingly considering activities that do not immediately include patients (such as hospital pharmacies, sterilization services, or inventory of medical and non-medical supplies) may represent a promising avenue for future research.  

\medskip

With regards to uncertainty modeling, the planning parameters that are considered as uncertain are so far mainly limited to durations\rev{, e.g., duration of surgery}. Notably, other very important issues such as no-show rates or availability of resources (staff, beds, etc.) are only rarely considered despite the substantial knock-on effects they can lead to in an integrated planning setting. No-shows of patients or other unexpected changes in patient demand might be particularly relevant here since the patient is usually the linking element between various steps along the care pathway~\cite{vanberkelSurvey}, so uncertain patient availability and demand will likely affect several parts of an integrated planning problem simultaneously. Staff unavailability, on the other hand, could become more and more important in the future due to the above-mentioned changes resulting from the \rev{ageing} population. Overall, we expect the uncertainty of parameters such as the number of available nurses, qualification levels of staff, the care chain (e.g., what resources are needed to treat a patient and are they actually available), and, finally, the flow of patients itself to be studied more in future research on integrated planning. Doing so could be especially valuable since the effects of these and other uncertain parameters \rev{as} well as knock-on effects caused by patient or resource unavailability can be much better understood with an integrated perspective (e.g., uncertain staff availability in the PACU can influence the number and types of surgeries that can be performed on a day and, therefore, also the number of beds required on the ICU or regular wards in the following days).
This seems particularly true for staff-related effects since staff typically \rev{work} in various places of a hospital (e.g., in the OT, the ICU, on wards, or in offices) and \rev{are} sometimes required to switch roles / positions during the day or week, which makes integrated planning seem inevitable in order to grasp the full consequences of uncertain staff availability and, as a result, ensure consistent availability of staff with the right expertise at the right place. 
\rev{Moreover, (medical) staff are often hard to replace both in the short and long term, which provides a further reason for considering uncertain staff availability as an especially relevant aspect that should be investigated more in future research.}

\medskip

\rev{Regarding the practical implementation of the research work on integrated planning problems in hospitals, an important aspect is for OR/MS to develop strong links not only to all involved personnel and hospital decision makers, but also to informatics~\cite{Erhard+etal:phys-sched-survey}. The latter is required to embed the developed approaches into hospital information systems in order to make them accessible for planners and decision makers as part of their daily practice.}

\rev{\subsection{Limitations}}

\rev{Like any literature review, our work also has limitations. First, since the focus of the review is on integrated planning in hospitals, integrated planning approaches outside of hospitals are not included in our review and only briefly discussed as an outlook in Section~\ref{sec:hospital+outside}. Moreover, the goal of providing a comprehensive review of the OR/MS literature related to integrated planning in hospitals without any restriction to specific resources or areas made it necessary to restrict the literature search to Operations Research \& Management Science journals and some additional relevant journals identified using domain knowledge. Even though this may have led to some relevant papers not being found by our initial search, the restriction of included journals was necessary since searching without it would have resulted in an unmanageable number of over 600.000 search results. We remark that the additional forward and backward search performed on the most cited papers from five different thematic areas was not restricted to specific journals and, thus, allowed us to identify additional relevant papers published in journals that were not included in the initial search. Here, the restriction of performing the forward and backward search only on a subset of the previously identified relevant papers was necessary due to the large number of relevant papers already found in our original search. Another limitation of our review is that, due to the broad scope of our review and the resulting very large number of relevant papers identified, in-depth discussion of individual papers could not be performed for all of these papers. While a detailed classification of all relevant papers with respect to all analyzed aspects can be found in the spreadsheet provided as an ancillary file, individual discussions of papers in the text had to be restricted to some of the most recent papers as presented in Section~\ref{sec:state-of-the-art} (State-of-the-art).
}

\medskip

\rev{\subsection{Take-home messages}}

We conclude our analyses of publications on integrated planning problems in hospitals with the following \emph{take-home messages:}
\begin{enumerate}
	\item \emph{The further planning problems move away from patients, the fewer integrated studies exist.} While patients are the main connecting element between resources and areas of a hospital to be integrated (and their pathways are often uncertain, too), there is still a lack of integrated studies that consider resources and activities that do not immediately include patients (e.g., sterilization, medical and non-medical supplies).
	\item \emph{Medical staff usually work in different places, which makes it even more important to consider integrated planning approaches.} Instead of following the patient through the hospital, movements of and requests for staff will be an interesting topic to follow.
	\item \emph{Knock-on effects (e.g., impacts of OT utilization on ward utilization) can only be fully understood if the system of interest is modeled in an integrated way.}  This, in turn, suggests that simulation studies (either stand-alone~\citesearchresults{Dosi_2021,Dwyer-Matzky_2021} or in connection with other planning approaches~\citesearchresults{Rachuba_2022,Oliveira_2022}) might receive even more interest in the future.
    \item \emph{\rev{Successful implementations of integrated planning approaches in practice are still rare.}} Despite the substantial share of papers that test their approaches in a case study, evidence of practical impact or successful implementation is still limited. This could be at least partly due to the increased amount of stakeholder involvement that might be required to implement an integrated approach in practice. While links to the involved personnel and decision makers are important for the successful implementation of any planning approach in a hospital, they seem particularly important for implementing integrated planning approaches that often involve multiple departments or decision making units of a hospital. According to our analysis of integrated planning approaches, it seems that simulation models seem to receive more buy-in from stakeholders so far compared to other approaches such as optimization models.
\end{enumerate}

\section*{Statements and Declarations}

\subsection*{Funding}
This research was funded by the Deutsche Forschungsgemeinschaft (DFG, German Research Foundation) -- Project number 443158418.

\rev{\subsection*{Author contributions}
\textbf{Sebastian Rachuba:} Conceptualization, Methodology, Validation, Formal Analysis, Investigation, Data Curation, Writing - Original Draft, Writing - Review and Editing, Visualization, Funding acquisition.
\textbf{Melanie Reuter-Oppermann:} Conceptualization, Methodology, Writing - Original Draft, Writing - Review and Editing, Funding acquisition.
\textbf{Clemens Thielen:} Conceptualization, Methodology, Validation, Formal Analysis, Data Curation, Writing - Original Draft, Writing - Review and Editing, Supervision, Project Administration, Funding acquisition.}

\subsection*{Competing Interests}
The authors have no relevant financial or non-financial interests to disclose.

\bibliographystyle{sn-basic}
\bibliography{references}

\begin{thebibliography}{56}
\bibitem[{Li et~al(2020)Li, Pan, and Xie}]{Li_2020}
Li N, Pan J, Xie X (2020) Operational decision making for a referral
  coordination alliance - when should patients be referred and where should
  they be referred to? Omega 96:102077
\end{thebibliography}

\nocitesearchresults{*}
\bibliographystylesearchresults{sn-basic}
\bibliographysearchresults{search-results_revised}

\end{document}